\documentclass[lettersize,journal]{IEEEtran}
\usepackage{amsmath,amsfonts}
\usepackage{algorithmic}
\usepackage{algorithm}
\usepackage{array}
\usepackage[caption=false,font=normalsize,labelfont=sf,textfont=sf]{subfig}
\usepackage{textcomp}
\usepackage{stfloats}
\usepackage{url}
\usepackage{verbatim}
\usepackage{graphicx}
\usepackage{siunitx}
\usepackage{gensymb}
\usepackage{cite}
\usepackage{booktabs}
\usepackage{multirow}
\usepackage{pifont} 
\usepackage[hidelinks]{hyperref} 

\usepackage{listings}
\usepackage{xcolor}

\newcommand{\newcontent}[1]{\textcolor{black}{#1}}

\begin{document}

\title{CLASH: Collaborative Large-Small Hierarchical Framework for Continuous Vision-and-Language Navigation}

\author{Liuyi Wang,~\IEEEmembership{Student Member,~IEEE},  Zongtao He, Jinlong Li, Ruihao Xia, Mengxian Hu, Chenpeng Yao, Chengju Liu, Yang Tang,~\IEEEmembership{Fellow,~IEEE}, Qijun Chen,~\IEEEmembership{Senior Member,~IEEE}
\thanks{Liuyi Wang, Zongtao He, Jinlong Li, Chenpeng Yao, Chengju Liu, and Qijun Chen are with Tongji University, Shanghai 201210, China. Ruihao Xia, and Yang Tang are with the Key Laboratory of Smart Manufacturing in Energy Chemical Process, Ministry of Education, East
China University of Science and Technology, Shanghai 200237, China. (Email: wly@tongji.edu.cn, qjchen@tongji.edu.cn).
This paper is supported by the National Natural Science Foundation of China under Grants (624B2105, 62473295, 62233013). (\textit{Corresponding Author: Qijun Chen.})
}
}



\maketitle

\begin{abstract}
Vision-and-Language Navigation in Continuous Environments (VLN-CE) requires robots to follow natural language instructions and navigate complex environments without prior maps. While recent vision-language large models demonstrate strong reasoning abilities, they often underperform task-specific panoramic small models in VLN tasks. To address this, we propose CLASH (Collaborative Large-Small Hierarchy), a VLN-CE framework that integrates a Reactive Small Model Planner (RSMP) with a Reflective Large Model Reasoner (RLMR). RSMP adopts a causal-learning-based dual-branch architecture to enhance generalization, while RLMR leverages panoramic visual prompting with chain-of-thought reasoning to support interpretable spatial understanding and navigation. We further introduce an Uncertainty-aware Collaboration Mechanism (UCM) that adaptively fuses decisions from both models. For obstacle avoidance, we replace the rule-based controller with a fully learnable point-goal policy in simulation. For real-world deployment, we design a LiDAR-based clustering module for generating navigable waypoints and pair it with an online SLAM-based local controller. CLASH achieves state-of-the-art (SoTA) results (\textit{ranking 1-st}) on the VLN-CE online leaderboard and relatively improves SR and SPL on the test set by 13.79\% and 13.73\% over the previous SoTA method. Real-world experiments further demonstrate CLASH’s strong robustness without any in-domain fine-tuning, validating its effectiveness in both simulation and deployment scenarios. The code is available at \href{https://crystalsixone.github.io/vln-clash.github.io/}{https://crystalsixone.github.io/vln-clash.github.io/}.
\end{abstract}


\setlength{\textfloatsep}{5pt}
\setlength{\abovecaptionskip}{5pt}
\setlength{\belowcaptionskip}{5pt}

\setlength{\abovedisplayskip}{3pt}
\setlength{\belowdisplayskip}{3pt}

\section{Introduction}
\label{sec_introduction}
\IEEEPARstart{G}{iven} a natural instruction such as \textit{``Go out of the current room and enter the kitchen on your right,”} Vision-and-Language Navigation (VLN)~\cite{anderson2018vision} tasks an embodied agent with interpreting natural-language instructions to navigate through complex, real-world environments—without access to prior maps.
Early VLN studies were conducted in discrete environments using the Matterport3D simulator~\cite{chang2017matterport3d}, where agents navigated among pre-defined panoramic viewpoints. To enhance real-world relevance, research has since shifted to continuous settings (VLN-CE)~\cite{krantz_beyond_2020}, where agents perceive egocentric views and execute low-level motor commands.
Over years of development, two dominant paradigms have emerged in VLN-CE: (1) end-to-end models that directly map raw visual and textual inputs to low-level actions~\cite{krantz_beyond_2020,he2024mee,zhang2024navid,cheng2024navila,wei2025streamvln}; and (2) hierarchical “brain-body” frameworks, where a high-level planner predicts navigation waypoints, which are then executed by a low-level controller~\cite{anderson2021sim,hong2022bridging,He2024_IAHWP,wang2023dreamwalker}. 
\newcontent{Initially, both paradigms relied on compact, task-specific small models, often enhanced by domain-specific data augmentation~\cite{wang2023scaling,wang2023res,wang2023pasts,li2022envedit}, improved multimodal semantic perception~\cite{wu2024tmm,tan2024self,wang2024seat,wang2023dual}, and better encoding of historical trajectories~\cite{hong2021vln,chen2021history,chen2022think,an2022bevbert}.}

\begin{figure}[t]
    \centering
    \includegraphics[width=\linewidth]{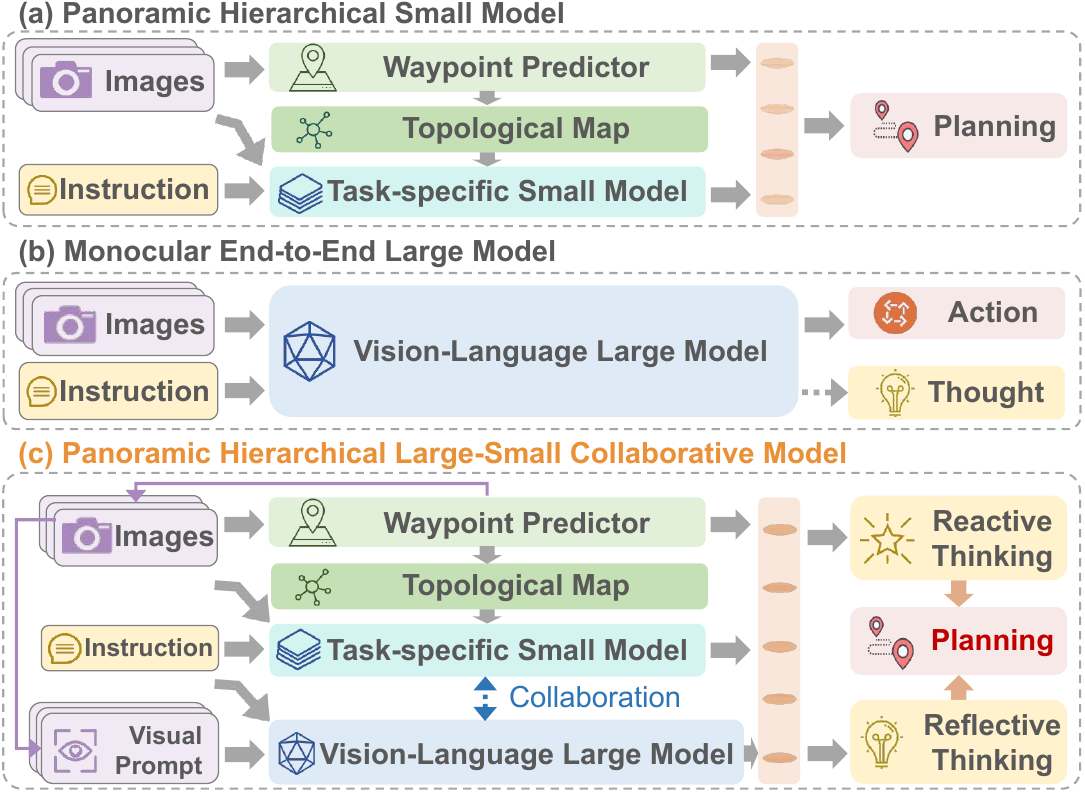}
    \caption{Comparison of CLASH with previous VLN-CE pipelines.}
    \label{fig:enter-label}
\end{figure}


\newcontent{Recently, multimodal large language models (MLLMs)~\cite{Qwen2.5-VL,wang2024qwen2,chen2024internvl}, pretrained on large-scale image–text datasets, have exhibited strong cross-modal reasoning abilities.} Their potential to generalize to embodied navigation has sparked growing interest in the VLN community.
\newcontent{Existing MLLM-based VLN methods also fall into two main paradigms. End-to-end methods (e.g., NaVid~\cite{zhang2024navid}, UniNaVid~\cite{zhang2024uni}, NaVILA~\cite{cheng2024navila}) fine-tune MLLMs to map monocular RGB observations directly to actions, treating visual inputs as continuous video streams. In contrast, hierarchical methods (e.g., NavGPT~\cite{zhou2024navgpt}, MapGPT~\cite{chen2024mapgpt}, InstructNav~\cite{long2024instructnav}, OpenVLN~\cite{qiao2024open}) employ MLLMs for high-level decision-making under zero- or few-shot settings.}

\newcontent{Our empirical comparisons across these paradigms reveal two consistent trends.}
(1) Hierarchical ``brain–body" frameworks with panoramic inputs consistently outperform fully end-to-end pipelines with monocular RGB inputs \newcontent{(e.g., g3D-LF~\cite{wang2025g3d} achieves 61\% SR, about 7\% higher than NaVILA~\cite{cheng2024navila}).}
(2) Within hierarchical frameworks, lightweight task-specific models still surpass fine-tuned MLLMs by a similar margin despite their smaller scale \newcontent{(e.g., MAGIC~\cite{wang2024magic} attains 79\% SR \textit{vs.} NavGPT2’s 72\% SR~\cite{zhou2024navgpt2})}.

Nevertheless, while small panoramic models perform well in simulation, they may overfit to specific scene styles, leading to degraded performance when transferred to more diverse environments~\cite{vlnpe}. Moreover, their opaque decision-making and limited interpretability hinder deployment in dynamic real-world scenarios. These observations raise a central question:
\textit{\textbf{How can we harness the complementary strengths of MLLMs and panoramic task-specific models to develop VLN agents that are both robust and generalizable?}}

To tackle these challenges, we introduce CLASH (Collaborative Large-Small Hierarchy), the first hierarchical VLN-CE framework that combines a Reactive Small-Model Planner (RSMP) with a Reflective Large-Model Reasoner (RLMR). \newcontent{As illustrated in Fig.~\ref{fig:enter-label}, this architecture leverages the complementary strengths of both models: the small model (under 2B parameters) using panoramic input serves as a specialized VLN expert delivering fast and reliable navigation decisions, while the MLLM (over 2B parameters) acts as a generalist expert offering broader commonsense reasoning and superior adaptability to dynamic or ambiguous environments.}
To maximize their individual capabilities, we enhance both components with targeted innovations. \newcontent{For RSMP, unlike existing VLN-CE methods~[32], [35], [36] that rely solely on global-level cross-modal alignment, we introduce a dual-branch framework that integrates: (i) fine-grained local cross-modal fusion to strengthen detailed environmental grounding and action decision-making, and (ii) causal learning that explicitly disentangles causalities from spurious correlations to improve generalization across unseen environments. For RLMR, we propose the panoramic visual prompting with chain-of-thought reasoning (PVP-CoT), which differs fundamentally from prior MLLM-based VLN approaches~[28], [29], [31] by providing spatially-aligned 360° context rather than fragmented separate views, thereby enhancing both interpretability and spatial reasoning capabilities.}

\newcontent{The critical question then becomes: \textit{how to effectively integrate RSMP and RLMR outputs for robust navigation?}} We address this through an uncertainty-aware collaboration mechanism (UCM) that adaptively fuses both models by estimating action confidence—prioritizing the small model when confident and the large model when uncertain. \newcontent{Unlike previous methods using cross-entropy~\cite{wang2024magic, qiao2024llm}, learned weights~\cite{liu2021efficient, NEURIPS2020_aab08546, feng2022uln} or multiple plausible outcomes~\cite{gal2016dropout,wang2024lookahead,koh2021pathdreamer} for uncertainty estimation, we adopt conformal prediction~\cite{shafer2008tutorial}, which provides statistically calibrated confidence measures with guaranteed coverage for more reliable decision-making. The advantages of using CP for uncertainty estimation are: (i) providing statistically reliable confidence measures that enable quantifiable control over decision reliability; (ii) not requiring model modification or additional training; and (iii) avoiding additional network prediction rounds, which would incur computational overhead.}

Beyond high-level reasoning, low-level execution is equally critical for bridging simulation and real-world deployment. \newcontent{Prior hierarchical VLN methods~\cite{an2024etpnav,wang2025g3d,wang2023gridmm} rely on simulated panoramic depth for waypoint prediction and use rule-based controllers that rotate randomly to escape collisions. While functional in simulation, this strategy suffers from two major drawbacks: (i) unsafe behavior, as random rotations and movements may cause collisions with obstacles; (ii) impracticality for real-world deployment, since consumer-grade panoramic cameras do not support pixel-aligned depth sensing required for accurate waypoint prediction.}

To address these challenges, we introduce a dual-level solution. \newcontent{For simulation, we replace rule-based control with a fully learnable point-goal navigation policy based on DDPPO~\cite{wijmansdd2019ddppo}, which demonstrates robust obstacle avoidance and deadlock recovery. For real-world deployment, where synchronized depth is unavailable and existing approaches~\cite{shi2025smartway,li2025ground} require inefficient multi-turn camera capture that severely disrupts navigation fluency, we propose a LiDAR-based clustering method for waypoint generation. This training-free approach applies K-means clustering over LiDAR cost maps to extract traversable candidates. For local execution,
we adopt an online SLAM-based controller that ensures robust localization and dynamic obstacle avoidance.}

Our major contributions are summarized as follows:

\newcontent{
\textbf{1) Novel Hybrid Architecture.} We introduce CLASH, the first VLN-CE framework that synergistically integrates a task-specific small model for reactive planning with an MLLM for reflective reasoning, combining task-specific efficiency with generalizable commonsense reasoning.}

\newcontent{
\textbf{2) Enhanced High-Level Reasoning.} We propose three key innovations: (i) a causal learning-based dual-branch framework that improves causal reasoning and generalization to unseen environments; (ii) panoramic visual prompting with chain-of-thought reasoning (PVP-CoT) for interpretable spatial reasoning; and (iii) an uncertainty-aware collaboration mechanism (UCM) that adaptively fuses decisions from both models based on confidence estimation.}

\newcontent{
\textbf{3) Practical Low-Level Execution.} We develop a dual-level solution bridging simulation and real-world deployment: a learnable DDPPO-based policy for simulation with robust obstacle avoidance, and a training-free LiDAR-based waypoint generation method with SLAM-based control for zero-shot real-world transfer.}

\newcontent{
\textbf{4) Superior Performance and Real-World Validation.} CLASH achieves state-of-the-art results on R2R-CE~\cite{krantz_beyond_2020} and REVERIE-CE~\cite{qi2020reverie}, ranking \textbf{\textit{1st}} on the VLN-CE leaderboard, with extensive real-world experiments validating strong sim-to-real transferability.
}

\section{Related Work}
\subsection{Vision-and-Language Navigation (VLN).}
The VLN task~\cite{anderson2018vision, krantz_beyond_2020,qi2020reverie,vlnpe} requires a robot to follow language instructions to navigate real-world environments. 
Early works expanded from single-camera to panoramic observations~\cite{fried2018speaker}, significantly improving performance. Subsequent efforts have explored diverse strategies, such as data augmentation~\cite{li2022envedit,wang2023pasts,source2025tan,wang2023scaling}, sensor fusion~\cite{wu2024tmm,wang2024seat}, and structured history modeling~\cite{yu2025moss,an2022bevbert}.
The continuous setting (VLN-CE)\cite{krantz_beyond_2020} requires agents to operate in a fine-grained 3D space using low-level actions. Within VLN-CE, research generally follows two main paradigms:

(a) Single-camera visual input:
Agents use first-person RGB or RGB-D views. Early methods employed cross-modal architectures with LSTMs for temporal modeling~\cite{He2025_MLANet,he2023learning,li2022reve-ce}. Recent work uses MLLMs for commonsense reasoning and generalization. While zero-shot performance is limited~\cite{wang2024qwen2,long2024instructnav}, fine-tuning with in-domain navigation data~\cite{zhang2024navid,cheng2024navila,zhang2024uni} significantly boosts action prediction, outperforming classical end-to-end monocular models. 

(b) Panoramic visual input: 
Another approach uses panoramic observations for 360-degree perception~\cite{anderson2021sim,hong2022bridging,He2024_IAHWP,wang2023dreamwalker}. \newcontent{Compared to single-camera settings, panoramic inputs can naturally provide richer environmental context, enhancing spatial awareness and decision-making.}
Panoramic inputs are typically used in hierarchical frameworks with high-level waypoint prediction and low-level action execution~\cite{an2024etpnav,krantz2021waypoint,shi2025smartway}. Some methods also integrate semantic mapping for exploration and trajectory reasoning~\cite{wang2025g3d,wang2023gridmm,an2022bevbert,wang2024lookahead}. 
\newcontent{However, most panoramic methods rely on task-specific small models. Recent efforts incorporating MLLMs in discrete panoramic settings~\cite{zhou2024navgpt,zhou2024navgpt2,chen2024mapgpt} still underperform specialized models. Additionally, VLN-Copilot~\cite{qiao2024llm} uses MLLMs to generate auxiliary guidance cues concatenated with navigation instructions to aid smaller models, but this approach demands extensive training data and ultimately depends on small models for inference. Although several recent studies~\cite{qiao2024open,long2024instructnav,shi2025smartway} explore zero-shot panoramic navigation with MLLMs, their prompt designs yield limited gains.}

This highlights a critical gap: how to effectively harness the strong world knowledge and reasoning abilities of MLLMs without sacrificing the robust task performance of small models. To address this, we propose a novel collaborative architecture that integrates both reactive small-model planning and reflective large-model reasoning, achieving improved overall performance and interpretability in VLN-CE tasks.

\subsection{\newcontent{Uncertainty Estimation in Reasoning and Navigation.}}
\newcontent{
Reliable uncertainty estimation plays a crucial role in enabling robust and safe decision-making for embodied reasoning and navigation agents. Some methods employ entropy-based estimation, which computes uncertainty from prediction entropy or cross-entropy values~\cite{wang2024magic,qiao2024llm}. While these measures are simple and efficient, they often lack statistical calibration, leading to overconfident predictions under distribution shift. Other methods leverage auxiliary-head or evidential modeling to learn additional network parameters for predicting confidence scores~\cite{liu2021efficient,NEURIPS2020_aab08546,feng2022uln}. Although capable of capturing both aleatoric and epistemic uncertainty, these methods require model modification, retraining, and ground-truth labels for supervision. Additionally, some methods~\cite{gal2016dropout,wang2024lookahead,koh2021pathdreamer} generate multiple plausible outcomes to represent uncertainty, but at the cost of significant computational overhead.}

\newcontent{
In contrast, we adopt conformal prediction (CP)~\cite{shafer2008tutorial,huang2024conformal} for uncertainty estimation. Unlike heuristic or architecture-dependent methods, CP provides statistically calibrated confidence measures with formal coverage guarantees, enabling principled control over decision reliability. Recent studies have demonstrated its effectiveness in robotic planning and LLM-based reasoning~\cite{renrobots}, where CP-guided confidence thresholds determine when an agent should trust its own plan or seek external guidance. Building upon this foundation, our approach leverages conformal prediction to achieve dynamic collaboration between the small and large models.}

\subsection{Multimodal Large Language Models in Embodied AI.}
Building on the remarkable success of large language models, recent efforts in Embodied AI have increasingly focused on developing vision-language foundation models to endow robots with more generalizable perception, reasoning, and decision-making capabilities. Broadly speaking, current approaches fall into three technical paradigms:
The first aims to build fully end-to-end MLLM-based systems that directly map raw observations to action sequences for task execution~\cite{kimopenvla,zitkovich2023rt,zhou2025chatvla}, striving for simplicity and task generalization.
The second adopts a hierarchical “brain-body” architecture, where a high-level MLLM (a cerebrum-like ``big brain”) interprets complex language commands and makes abstract decisions, which are then executed by a low-level controller (a cerebellum-like “little brain”)~\cite{driess2023palm,intelligence2025pi,huang2024rekep,belkhale2024rth}.
The third follows a multi-agent collaborative design, where a complex task is decomposed into smaller subtasks, each handled by an expert agent. Their outputs are then coordinated to guide the final execution~\cite{li2023camel,guo2024embodied,feng2025multi}.

\section{Vision-Language Navigation Definition}

We formulate VLN in a continuous 3D environment, where an embodied agent follows a natural language instruction $I$ to reach a goal. The agent operates under low-level controls: \texttt{FORWARD}, \texttt{TURN-LEFT}, \texttt{TURN-RIGHT}, and \texttt{STOP}. At each timestep $t$, the agent observes a panoramic view $E$, divided into 12 horizontal sub-images at $30\degree$ intervals. Each view has a resolution of $256 \times 256$ pixels and a $75\degree$ vertical field of view. The agent’s orientation is represented as $\gamma = (\sin\theta, \cos\theta, \sin\phi, \cos\phi)$, where $\theta$ and $\phi$ denote heading and elevation.
In continuous settings, executing a selected waypoint requires a low-level controller. Success is defined as issuing a \texttt{STOP} within 3 meters of the goal location.

\begin{figure*}[thb]
    \centering
    \includegraphics[width=\linewidth]{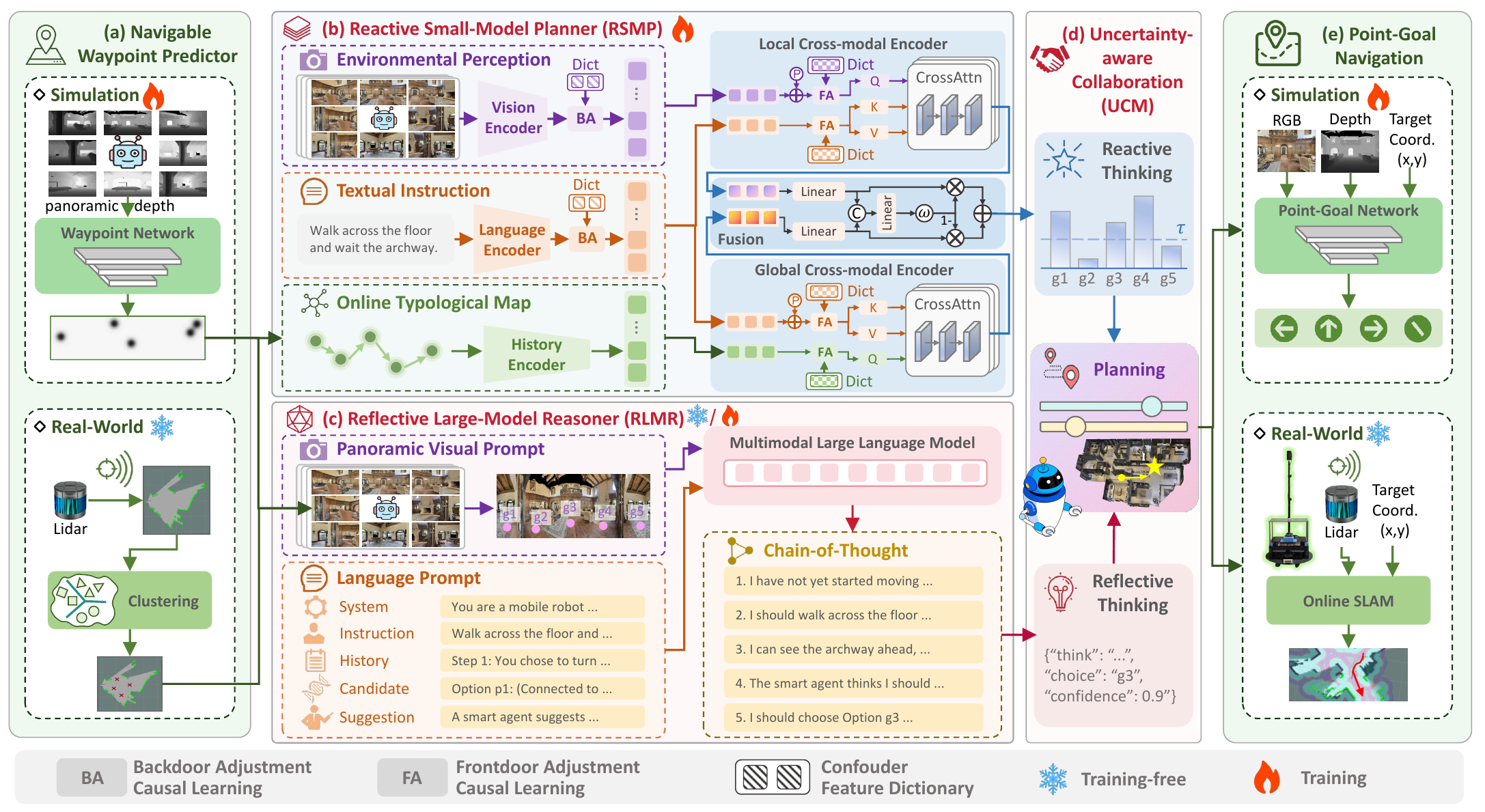}
    \caption{Overview of the CLASH framework. First, the model employs (a) a navigable waypoint predictor to generate candidate waypoints in the surrounding environment. It then produces action predictions using (b) RSMP and (c) RLMR. When the estimated uncertainty is high, the (d) uncertainty-aware collaboration mechanism (UCM) is activated to fuse the decisions of the two models. Finally, low-level navigation is executed via (e) a Point-Goal Navigation module.}
    \label{fig_overall}
\end{figure*}

\section{High-level Decision-Making Module}

\subsection{Reactive Small-Model Planner (RSMP)}
\label{sec_RSMP}
VLN is a sequential decision-making problem that can be formulated as a partially observable Markov decision process (POMDP). It is essential to effectively represent and align visual, linguistic, and memory features. To improve robustness beyond purely observational learning, we first incorporate causal learning~\cite{wang2024causal}, which helps decouple spurious dependencies and enhance generalization. In addition, since existing hierarchical VLN-CE methods~\cite{an2024etpnav,wang2025g3d,wang2024lookahead} mainly fuse the global topological graph with the instruction and often overlook fine-grained local cues, we adopt a dual-graph structure following prior work in the discrete setting~\cite{chen2022think,wang2023scaling,wang2023dual}.

\subsubsection{\newcontent{Causal Learning}}
\label{subsubsec_causal_learning}
\newcontent{Generalization to unseen environments is crucial for VLN agents. However, conventional training often leads models to rely on spurious correlations in observational data, which may limit robustness under distribution shifts. Causal learning~\cite{pearl2018book,wang2024causal} offers a principled framework for reasoning beyond such correlations. Specifically, traditional methods model the observational likelihood $P(\mathcal{Y}|\mathcal{X})$, where $\mathcal{X}$ denotes multi-modal inputs (vision, language, and history) and $\mathcal{Y}$ represents the action prediction. This formulation does not explicitly account for \textit{confounders}—latent factors that jointly influence inputs and outputs and may introduce shortcut dependencies.}
\newcontent{Causal inference introduces the \textit{do}-operator to model interventional reasoning and decouple such entangled effects. Following~\cite{wang2024causal}, confounders are categorized into two types: observable confounders (\textit{e.g.}, room-type cues in vision, directional keywords in language) and unobservable confounders (\textit{e.g.}, stylistic variations, linguistic patterns, trajectory tendencies). To better capture their respective influences in VLN-CE, we incorporate back-door and front-door adjustment modules into the RSMP.}

\newcontent{Back-door Adjustment (BA) is used for observable confounders $\mathcal{Z}^o$. By stratifying over confounder categories and averaging their effects, the causal effect is estimated as:}
\begin{equation}
\newcontent{P(\mathcal{Y}|\text{do}(\mathcal{X})) = \sum_{z^o} P(\mathcal{Y}|\mathcal{X},z^o)P(z^o),}
\end{equation}
\newcontent{where $P(z^o)$ is the marginal distribution of confounders. Confounder feature dictionaries (e.g., room-type features, keyword embeddings) are pre-constructed and used to approximate the expected adjustment term. This process can be formulated as $BA(\mathbf{x},\mathbf{z}) = f_x(\mathbf{x})+\mathbb{E}_z[f_z(\mathbf{z})]$, where bold symbols represent feature vectors and $f_x$, $f_z$ are the mapping functions.}

\newcontent{Front-door Adjustment (FA) addresses unobservable confounders $\mathcal{Z}^u$ by introducing a mediator $\mathcal{M}$ that lies on the causal path $\mathcal{X} \rightarrow \mathcal{M} \rightarrow \mathcal{Y}$. The causal effect is derived as:}
\begin{equation}
  \newcontent{P(\mathcal{Y}|\text{do}(\mathcal{X})) = \sum_{m}\sum_{x'} P(\mathcal{Y}|m,x')P(x')P(m|\mathcal{X}),}
\end{equation}
\newcontent{where $x'$ represents potential input samples from the entire training feature space, and $m$ denotes mediator features. This adjustment is approximated through cross-sampling from global feature dictionaries and in-sampling from current inputs, implemented efficiently via multi-head attention mechanisms. This process can be formulated as $FA(\mathbf{x},\mathbf{x}')=\mathbb{E}_{x'}[\mathbf{x}']+\mathbb{E}_{m|x}[\mathbf{m}].$}
\newcontent{By integrating both BA and FA into the network architecture, the model learns more reliable feature representations across modalities, enhancing generalization to unseen environments~\cite{wang2024causal}.}

\subsubsection{Framework of the RSMP}
\label{subsec_rsmp_framework}
Specifically, the key components of RSMP are summarized below:

\textbf{a. Language Encoding Module}:
To capture the semantic structure of the instruction $I$, a pretrained language model such as BERT~\cite{devlin2019bert} is utilized. Let $\psi(\cdot)$ be the linear transformation layer, $\phi(\cdot)$ the embedding layer, and $\mathcal{F}_I(\cdot)$ the transformer encoder. The resulting language feature representation $Q_s \in \mathbb{R}^{L \times d_h}$ is computed as $Q_s = \mathcal{F}_I(\psi_i(I) + \phi_i(P_I))$,
\newcontent{where $L$ denotes the number of the instruction tokens, $d_h$ is the hidden dimension of the feature embeddings}, and $P_I$ denotes the absolute positional indices of the tokens. 
\newcontent{When the causal learning module is applied, the text content causal representations become $G_s=LN[\psi_s(BA(Q_s, Z_k))]$, where $LN$ denotes the layernorm function, and $Z_k$ is the textual confounder features.}

\textbf{b. Panoramic Visual Encoding Module}:
To comprehensively perceive its surroundings, the agent processes a panoramic observation $E$ composed of 12 perspective sub-images. Each sub-image is first encoded by the pre-trained visual encoder $\mathcal{F}_V(\cdot)$ (like CLIP~\cite{radford2021learning}). The relative orientation of each sub-image is represented by a directional embedding $\gamma_{im} = {(\sin \theta_i, \cos \theta_i, \sin \beta_i, \cos \beta_i)}_{i=1}^{12}$. A multi-layer self-attention encoder ($\mathcal{F}_{\text{SA}}$) then fuses the multi-view features to enable cross-view interaction, obtaining $Q_v\in \mathbb{R}^{12\times d_h}$:
\begin{align}
    V_v &= \psi_v(\mathcal{F}_V(E)), \\
    Q_v &= \mathcal{F}_{\text{SA}}(V_v + \phi_d(\gamma_{im})).
\end{align}
\newcontent{Similarly, when the causal learning is adopted, the visual causal features become $\widetilde{Q}_v=LN[\psi_r(BA(Q_v, Z_r))]$, where $Z_r$ denotes the visual confounder features.}

\textbf{c. Global-Local Cross-Modal Localization Module}:
\newcontent{In widely used VLN-CE hierarchical frameworks~\cite{an2024etpnav,wang2025g3d,wang2024lookahead}, cross-modal fusion is typically performed only at the global level between the instruction and the historical topology representation. However, this design may overlook fine-grained local context. Motivated by the effectiveness of dual local–global architectures in discrete settings~\cite{chen2022think,wang2023scaling}, we introduce a local cross-modal fusion branch to enhance detailed environmental grounding and action-level decision-making.}

At each timestep, the historical feature $\bar{Q}_v$, which is compressed from the panoramic embedding, is stored in a topological map. A \texttt{[CLS]} token is prepended to both local ($G_v \in \mathbb{R}^{L_l\times d_h}$) and global ($G_h \in \mathbb{R}^{L_g \times d_h}$) visual sequences to facilitate the \textit{stop} action. \newcontent{$L_l$ and $L_g$ denote the length of the local and global visual sequences, respectively.} 
\newcontent{To introduce the frontdoor causal learning module, the presentations for instruction, vision, and memory become: $R_s = FA(G_s, Z_{us}),\, R_v = FA(G_v, Z_{uv}),\, R_h = FA(G_h, Z_{uh}),$ where $Z_{ux}$ denote the corresponding confounder features.}
Then, two cross-modal attention encoders ($\mathcal{F}_{\text{CA}}$) integrate visual and historical queries with instruction embeddings:
\begin{align}
F_l &= \mathcal{F}_{\text{CA}}(R_v, R_s, R_s), \\
F_g &=\mathcal{F}_{\text{CA}}(R_h, R_s, R_s).
\end{align}

\textbf{d. Planning Decision Module}:
Local visual features are first aligned with global historical features according to the corresponding candidate points. A learnable weight $w$ is used to fuse local and global predictions adaptively. The combined logits $B_f$ are passed through a softmax layer to yield the final action distribution over candidate waypoints:
\begin{align}
B_f &= wF_l + (1 - w)F_g, \\
b_i &= \frac{\exp(B_{f,i})}{\sum_j \exp(B_{f,j})}, \label{eq_softmax}
\end{align}
\newcontent{where $B_{f,i}$ denotes the $i$-th element of the fused logit vector $B_f$, and $b_i \in [0, 1]$ represents the predicted probability for the $i$-th waypoint candidate.} During inference, the agent selects the candidate node with the highest probability, i.e., $\arg\max_i b_i$.

\subsection{Reflective Large-Model Reasoner (RLMR)}
\label{sec_RLMR}
Despite strong benchmark performance, task-specific small models still lack open-set reasoning ability and process interpretability when faced with diverse real-world scene styles and regions. Therefore, we introduce an MLLM as a reflective reasoner to enhance both generalization and transparency.

\subsubsection{Panoramic Visual Prompt (PVP)}
\begin{figure}
    \centering
    \includegraphics[width=\linewidth]{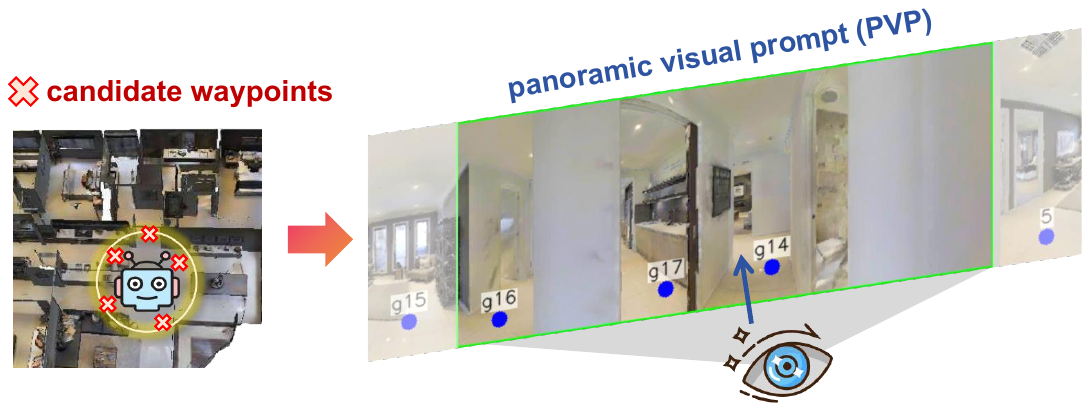}
    \caption{\newcontent{Illustration of PVP.} Nodes prefixed with “\textit{g}” are ghost candidates, and numbered markers indicate visited locations.}
    \label{fig_pwvp}
\end{figure}

To enable holistic spatial understanding, different from prior methods that rely on discrete sub-views or sequential video streams~\cite{chen2024mapgpt,zhou2024navgpt2,cheng2024navila}, we directly feed the full panoramic observation into the MLLM. Candidate waypoints are projected onto the panorama to visually encode their spatial positions, thereby enhancing the alignment between language instructions and navigational intent.
\newcontent{Additionally, we observe that the lateral edges of the equirectangular panorama correspond to the agent’s backward view, which can mislead MLLMs into interpreting rear-view regions as lateral surroundings. Since navigation reasoning primarily depends on the forward field of view, we introduce a semi-transparent mask that attenuates the visual saliency of rear-view regions while preserving global scene continuity. This design reduces distractive cues from the backward view and helps the model concentrate on task-relevant front-view information.
Our design is theoretically motivated by recent advances in attention prompting~\cite{yu2024attention}, which demonstrate that explicit visual prompts or region-wise masking can steer the attention distribution of large vision-language models and improve spatial reasoning.}
An illustration of PVP and the masking effect is shown in Fig.~\ref{fig_pwvp}.

\subsubsection{Structural vision-language prompt}
\begin{figure*}
    \centering
    \includegraphics[width=\linewidth]{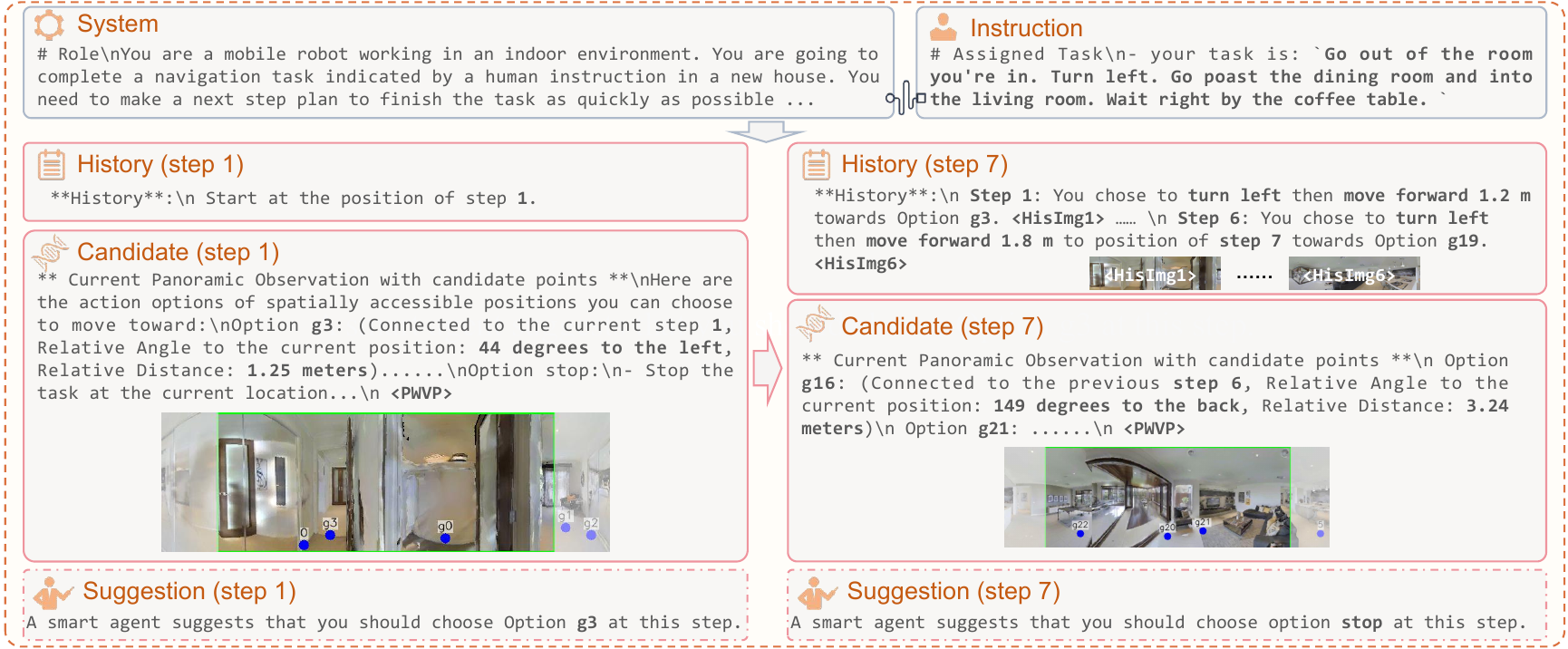}
    \caption{Overview of the structural vision-language prompt for VLN. It consists of five key components: system, instruction, history, candidate, and suggestion.}
    \label{fig:vision-language-prompt}
\end{figure*}
As shown in Fig.~\ref{fig:vision-language-prompt}, the Structural input prompt consists of five components:
 (a) System: Briefly defines the agent's role, input, and expected output.
 (b) Instruction: The natural language command to follow.
 (c) History: Past actions with relative directions, distances, and panoramic views.
 (d) Candidate: Descriptions of current navigable waypoints, including connectivity, directions, and distances. 
 (e) Suggestion: Action recommendation from the small model to aid decision-making.

\subsubsection{Causal chain-of-thought}

To enhance reasoning and interpretability, we apply chain-of-thought (CoT) prompting~\cite{wei2022chain} to guide the RLMR to reason through five steps: trajectory status, action planning, visual grounding, suggestion evaluation, and final decision, with an assigned confidence score $\mathbf{p}^{\text{RLMR}}$. The complete prompt template is provided in the appendix.

\subsection{Uncertainty-Aware Collaboration Mechanism (UCM)}
\label{subsec_ucm}
Existing works demonstrated that RSMP, operating within a panoramic-view hierarchical structure, consistently outperforms RLMR, even when the latter is fine-tuned on the in-domain VLN dataset~\cite{wang2024magic, zhou2024navgpt, zhou2024navgpt2}. While RLMR has the great potential to offer greater transparency and reflective reasoning, it is more prone to errors in long-horizon embodied reasoning tasks. To effectively integrate the strengths of both agents, we propose an Uncertainty-Aware Collaboration Mechanism (UCM), which leverages RSMP's uncertainty estimates to adaptively mediate decision fusion with RLMR.

\subsubsection{\newcontent{Conformal Prediction for Uncertainty Quantification}}
\label{subsubsec_conformal_prediction}

To quantify RSMP's prediction uncertainty, \newcontent{we adopt conformal prediction (CP)~\cite{shafer2008tutorial,huang2024conformal}, a distribution-free framework that provides statistically calibrated confidence measures with formal coverage guarantees. CP constructs a prediction set $\mathcal{C}(\mathcal{X}) \subseteq \mathcal{Y}$ such that, in expectation over the data distribution, the true label $\hat{y}$ lies within $\mathcal{C}(\mathcal{X})$ with probability at least $1 - \epsilon$. This coverage guarantee is statistically derived from the calibration set and is not enforced during evaluation or test inference. Specifically, $\mathcal{X}$ denotes the input observation space (comprising visual observations, language instructions, and navigation history), $\mathcal{C}(\cdot)$ denotes a set of possible output candidates, $\mathcal{Y}$ represents the action label space (possible waypoint candidates), and $\epsilon \in [0, 1]$ is a user-specified miscoverage rate (i.e., the allowed probability of the prediction set not containing the true action).}

\newcontent{\textbf{Calibration Phase:}} \newcontent{We randomly sample 50\% of the training data as a calibration set $\{(x_i, y_i)\}_{i=1}^n$}, where $n$ denotes the number of calibration samples, $x_i$ is the $i$-th input observation, and $y_i$ is the corresponding ground-truth action. \newcontent{A nonconformity score is defined as $s_i = S(x_i, y_i)$ for each example $(x_i, y_i)$ in the calibration set. This score measures the degree of deviation between the given example and the training data. Higher scores indicate lower confidence in the correct action. Then, $S(x_i, y_i)$ is computed as:}
\begin{equation}
\newcontent{S(x_i, y_i) = 1 - b_i(x_i, y_i),}
\end{equation}
\newcontent{where $b_i(x_i, y_i)$ computed by eq.\eqref{eq_softmax} denotes the predicted probability assigned by RSMP to the ground-truth action $y_i$ for input $x_i$.} Then the $(1 - \epsilon)$-quantile of the calibration scores $\{s_i\}_{i=1}^n$ is computed to obtain the conformity threshold $\tau$:
\begin{equation}
\newcontent{\tau = \text{Quantile}_{1-\epsilon}(\{s_i\}_{i=1}^n).}
\end{equation}

\newcontent{\textbf{Inference Phase:}} At test time, given a new observation $x$, we construct the prediction set by including all candidate actions $y \in \mathcal{Y}$ whose nonconformity score $S(x, y) \leq \tau$:
\begin{equation}
\mathcal{C}_{1-\epsilon}(x;\tau) := \{y \in \mathcal{Y} : S(x, y) \leq \tau\}.
\end{equation}
\newcontent{The cardinality of the prediction set, denoted as $|\mathcal{C}(x)|$, naturally reflects the model's uncertainty: a larger set indicates higher ambiguity among multiple plausible actions, while a singleton set ($|\mathcal{C}(x)| = 1$) implies high confidence. }

\subsubsection{\newcontent{Adaptive Decision Fusion}}
\label{subsubsec_adaptive_decision_fusion}

Building upon the conformal prediction set $\mathcal{C}(x)$, UCM performs adaptive decision fusion for both models. \newcontent{The complete workflow is provided in Algorithm~\ref{alg:ucm}.} 
When $|\mathcal{C}(x)| = 1$, RSMP is confident in its prediction and the selected action is directly executed. 
When $|\mathcal{C}(\mathcal{X})| > 1$, RSMP exhibits uncertainty over multiple plausible candidates. In this case, the prediction set $\mathcal{C}(\mathcal{X})$ is forwarded to RLMR for reflective reasoning. 
If RSMP and RLMR produce conflicting decisions, we resolve the disagreement by fusing their outputs according to the uncertainty distribution. Let $\mathbf{p}^{\text{RSMP}} \in \mathbb{R}^K$ denote RSMP's action probability distribution over $K$ waypoint candidates.
Let $L_{\text{max}} = 10$ denote the normalization factor. The uncertainty of RSMP is then defined as:
\begin{equation}
\alpha = \min\left(\frac{|\mathcal{C}(\mathcal{X})|}{L_{\text{max}}},\ 0.9\right),
\end{equation}
where $\alpha \in [0, 0.9]$ is used to quantify the uncertainty level. A higher $\alpha$ indicates greater uncertainty.
\newcontent{RLMR generates a reflective decision with confidence score $c^{\text{RLMR}}$, which is converted to a sparse distribution $\mathbf{p}^{\text{RLMR}} \in \mathbb{R}^K$ by assigning $c^{\text{RLMR}}$ to the selected action and 0 elsewhere. This one-hot-like representation preserves RLMR's interpretability while enabling probabilistic fusion.}
The final fused action distribution is computed as:
\begin{equation}
\mathbf{p}^{\text{UCM}} = (1 - \alpha) \cdot \mathbf{p}^{\text{RSMP}} + \alpha \cdot \mathbf{p}^{\text{RLMR}}.
\end{equation}

The action selected by the agent is then given by $a^* = \arg\max_k \ p_k^{\text{UCM}}$.
This strategy prioritizes RSMP’s decisions when confident, while shifting weight to RLMR’s reflective reasoning under uncertainty, enabling more robust and reliable navigation through complementary strengths.

\begin{algorithm}[t]
\caption{\newcontent{Uncertainty-Aware Collaboration Mechanism}}
\label{alg:ucm}
\begin{algorithmic}[1]
    \REQUIRE Input instance $x = (E, I, H)$ where $E, I$ and $H$ denote observation, instruction, and history; waypoint candidates $\mathcal{W} = \{w_k\}_{k=1}^K$; conformal threshold $\tau$.
    \ENSURE Selected action $a^*$
    \STATE // \textit{RSMP Prediction \& Conformal Set Construction}
    \STATE Get action distribution: $\mathbf{p}^{\text{RSMP}} \leftarrow \text{RSMP}(E, I, H, \mathcal{W})$ \hfill
    \STATE Compute nonconformity scores: $s_k = 1 - p_k^{\text{RSMP}}$ for $k = 1, \ldots, K$
    \STATE Construct prediction set: $\mathcal{C}(x) = \{w_k : s_k \leq \tau\}$
    \STATE // \textit{Candidate Filtering \& RLMR Invocation}
    \IF{$|\mathcal{C}(x)| = 1$}
        \STATE $a^* \leftarrow$ the single element in $\mathcal{C}(x)$ \hfill // High confidence
        \RETURN $a^*$
    \ELSE
        \STATE // Multiple plausible candidates, invoke RLMR
        \STATE $\langle a^{\text{RLMR}}, c^{\text{RLMR}} \rangle \leftarrow \text{RLMR}(x, I, H, \mathcal{C}(x))$
        \STATE Initialize $\mathbf{p}^{\text{RLMR}} \in \mathbb{R}^K$ as zero vector
        \STATE Get sparse action distribution $p_{a^{\text{RLMR}}}^{\text{RLMR}} \leftarrow c^{\text{RLMR}}$ \hfill
    \ENDIF
    \STATE // \textit{Adaptive Fusion}
    \STATE Compute uncertainty weight $\alpha$
    \STATE Fuse distributions: $\mathbf{p}^{\text{UCM}} = (1 - \alpha) \cdot \mathbf{p}^{\text{RSMP}} + \alpha \cdot \mathbf{p}^{\text{RLMR}}$
    \STATE Select final action $a^* \leftarrow \arg\max_{k} p_k^{\text{UCM}}$ \hfill
    \RETURN $a^*$
\end{algorithmic}
\end{algorithm}

\section{Low-Level Execution Module}
\label{sec_low_level_exe}
While often overlooked, low-level execution is critical for VLN-CE, especially given the sensor disparities between simulators and real-world platforms. We develop separate implementations of the waypoint and point-goal modules to ensure reliable and adaptable control in both settings.

\subsection{Waypoint Predictor (WP)}
\label{subsec_wg}

\subsubsection{Panoramic-RGBD end-to-end waypoint predictor}
\label{subsubsec_end2end_wp}
Most prior hierarchical VLN methods~\cite{krantz2022sim, an2024etpnav} adopt an end-to-end waypoint predictor. The model takes RGB $\{v_i\}_{i=1}^{12}$ and depth $\{d_i\}_{i=1}^{12}$ panoramas, and encodes them using pre-trained visual backbones (denoted as $VE(\cdot)$ and $DE(\cdot)$, respectively). The visual embeddings are then processed by a Transformer encoder ($\mathcal{F}_{\text{SA}}$). A classifier ($\phi_c(\cdot)$) outputs a heatmap over a discretized 2D space defined by 120 angles (at $3\degree$ resolution) and 12 distances (from 0.25m to 3.0m in 0.25m increments). Finally, non-maximum suppression (NMS) is applied to the heatmap to extract the top-P candidate waypoints.
\begin{align}
    R_v &= VE(\{v_{i}\}), \, R_d = DE(\{d_i\}), \\
    W_r &= \mathcal{F}_\text{SA}(Concat(R_v;R_d)), \\
    W_o &= NWS(\phi_{c}(W_r)).
\end{align}

While effective in simulation, this approach faces real-world challenges. Panoramic cameras often lack synchronized RGB-D data, unlike simulators with perfect depth, leading to poor transferability. Moreover, with LiDAR or range sensors, key waypoints can often be identified directly, reducing the need for complex end-to-end models.

\subsubsection{Lidar-based clustering waypoint predictor}
\label{subsubsec_lidar_based_waypoint}
\begin{figure}
  \centering
  \includegraphics[width=\linewidth]{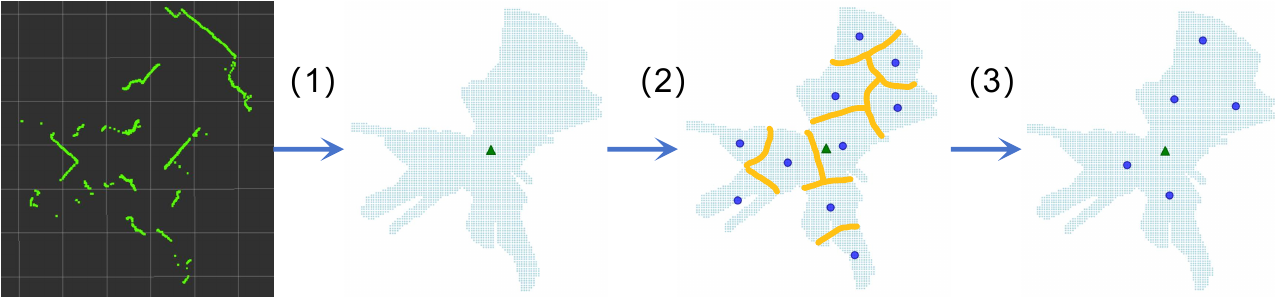}
  \caption{\newcontent{Pipeline of the waypoint generation. (1) Generates navigable points from LiDAR scans. (2) Clusters the navigable points. (3) Refines the cluster centers by removing points too close to obstacles or each other.}}
  \label{fig:lidar_clustering}
\end{figure}

We introduce a training-free, LiDAR-based method for efficient waypoint generation via clustering (Algorithm~\ref{alg:lidar_clustering}, Fig.~\ref{fig:lidar_clustering}). A 2D occupancy cost map is first constructed from LiDAR scans, encoding obstacle proximity to indicate traversability. Navigable regions are extracted, and K-means clustering identifies $K$ cluster centers as waypoint candidates. These are refined by removing points too close to obstacles or each other. Waypoint coordinates are recorded in both the robot's local and global frames using the ROS TF tree. By default, we use $K=10$ clusters with a $120 \times 120$ grid at $0.05$m resolution.
This approach requires no training or annotations, offering high data efficiency and low deployment cost. It also leverages LiDAR’s robustness, making it well-suited for real-world applications.

\begin{algorithm}[t]
\caption{LiDAR-Based Clustering Waypoint Predictor}
\label{alg:lidar_clustering}
\begin{algorithmic}[1]
    \REQUIRE LiDAR scan $Ld$, number of clusters $K$
    \ENSURE Waypoint candidates $\mathcal{W} = \{w_1, \ldots, w_K\}$
    \STATE Generate 2D occupancy cost map $\mathcal{M}$ from $Ld$
    \STATE Extract navigable points $\mathcal{P}$ from $\mathcal{M}$
    \STATE Apply K-means to $\mathcal{P}$ to obtain $K$ cluster centers
    \STATE Filter out centers near obstacles or each other
    \STATE Transform cluster centers to robot and world frames
    \RETURN $\mathcal{W}$
\end{algorithmic}
\end{algorithm}

\subsection{Local Navigation Controller (LNC)}

\subsubsection{RGB-D-based point-Goal navigation network}
For the local navigation controller, ETPNav~\cite{an2024etpnav} uses a ``rotate-then-forward" strategy with a \textit{Tryout} heuristic: if repeated \texttt{FORWARD} actions fail to change position, the agent rotates through preset angles $\Delta \Theta^{try}$ and retries.
While effective in simulation, this trial-and-error approach is unsafe for real-world use. It often fails in scenarios with elevation changes, such as approaching a target atop stairs from the side. To address this, we adopt DDPPO~\cite{wijmansdd2019ddppo}, an end-to-end policy that maps RGB-D inputs and goals to actions, improving obstacle avoidance and deadlock handling in simulation.

\subsubsection{SLAM-based point-goal navigation algorithm}
\label{subsubsec_slam_pointGoal}
LiDAR-based SLAM systems are widely used for robust real-world navigation, offering mapping, localization, and path planning in unknown environments without pre-built maps.
Although DDPPO performs well in simulation, prior work~\cite{gervet2023navigating} shows limited transferability to the real world. To ensure reliable physical deployment, we adopt an online SLAM-based pipeline using ROS \texttt{move\_base} for waypoint navigation. This hybrid approach bridges the sim-to-real gap and ensures safe, efficient traversal.

\section{Experiment}
\label{sec_experiment}
\subsection{Experimental Settings}
\subsubsection{Datasets}
The proposed method is evaluated on two widely used VLN-CE benchmarks: R2R-CE and REVERIE-CE, which are adapted from the discrete-path datasets~\cite{anderson2018vision,qi2020reverie}. 
The R2R-CE dataset consists of real-world scanned environments across 90 buildings, with an average length of 32 words. It is split into \texttt{train} (10,819 pairs, 61 scenes), \texttt{val seen} (same buildings as training, 778 pairs and 61 scenes), \texttt{val unseen} (different buildings, 1,839 pairs and 11 scenes), and \texttt{test unseen} (3,408 pairs and 18 scenes). The \texttt{test} set is evaluated via an online server for fair benchmarking. The REVERIE-CE dataset includes 10,318 panoramas from 86 buildings and 21,702 crowd-sourced instructions, averaging 18 words per instruction.

\begin{table*}[t]
\caption{Comparison with other SoTA methods on the R2R-CE dataset~\cite{krantz_beyond_2020}. `-' indicates the unavailable statistics. `Pano' denotes whether panoramic visual input is used. $\dagger$ means the usage of the additional data beyond the VLN.}
\label{tab_r2rce}
\resizebox{\linewidth}{!}{
\begin{tabular}{@{}c|l|c|ccccc|ccccc|ccccc@{}}
\toprule
\multirow{2}{*}{Type} &
  \multirow{2}{*}{Method} &
  \multirow{2}{*}{Pano} &
  \multicolumn{5}{c|}{Val Seen} &
  \multicolumn{5}{c|}{Val Unseen} &
  \multicolumn{5}{c}{Test Unseen} \\
                      & &            & TL    & NE$\downarrow$   & OSR$\uparrow$  & SR$\uparrow$ & SPL$\uparrow$ & TL    & NE$\downarrow$   & OSR$\uparrow$ & SR$\uparrow$ & SPL$\uparrow$ & TL    & NE$\downarrow$   & OSR$\uparrow$ & SR$\uparrow$ & SPL$\uparrow$ \\ \midrule
\multirow{15}{*}{\uppercase\expandafter{\romannumeral1}}   & CM2~\cite{georgakis2022cross}       & \ding{55} & 12.05 & 6.10 & 51   & 43 & 35  & 11.54 & 7.02 & 42  & 34 & 28  & 13.90 & 7.70 & 39  & 31 & 24  \\
                      & CWP-CMA~\cite{hong2022bridging}  & \ding{51}  & 11.47 & 5.20 & 61   & 51 & 45  & 10.90 & 6.20 & 52  & 41 & 36  & 11.85 & 6.30 & 49  & 38 & 33  \\
                      & CWP-RecBERT~\cite{hong2022bridging} & \ding{51} & 12.50 & 5.02 & 59   & 50 & 44  & 12.23 & 5.74 & 53  & 44 & 39  & 13.51 & 5.89 & 51  & 42 & 36  \\
                      & GridMM~\cite{wang2023gridmm}    & \ding{51} & 12.69 & 4.21 & 69   & 59 & 51  & 13.36 & 5.40 & 57  & 50 & 46  & 13.31 & 5.55 & 57  & 49 & 45  \\
                      & DreamWalker~\cite{wang2023dreamwalker} & \ding{51} & 11.60 & 4.09 & 66   & 59 & 48  & 11.30 & 5.53 & 59  & 49 & 44  & 11.80 & 5.48 & 57  & 49 & 44  \\
                      & ScaleVLN~\cite{wang2023scaling}   & \ding{51}  & -     & - & -   & - & -  & -     & 4.80 &  -  & 55 & 51  & -     & 5.11 & -  & 55 & 50  \\
                      & ETPNav~\cite{an2024etpnav}   & \ding{51}  & 11.78 & 3.95 & 72   & 66 & 59  & 11.99 & 4.71 & 65  & 57 & 49  & 12.87 & 5.12 & 63  & 55 & 48  \\
                      & BEVBert~\cite{an2022bevbert}  & \ding{51}  & -     & -    & -    & -  & -   & -     & 4.57 & 67  & 59 & 50  & -     & 4.70 & 67  & 59 & 50  \\
                      & HNR~\cite{wang2024lookahead}     &  \ding{51}  & 11.79 & 3.67 & 76   & 69 & 61  & 12.64 & 4.42 & 67  & 61 & 51  & 13.03 & 4.81 & 67  & 58 & 50  \\
                      & UnitedVLN~\cite{dai2024unitedvln} & \ding{51} & -     & 3.30 & 78   & 70 & 62  & -     & 4.29 & 70  & 62 & 51  & -     & 4.69 & 68  & 59 & 49  \\ 
                      & g3D-LF~\cite{wang2025g3d} &  \ding{51} & - & - & - & - & - & - & 4.53 & 68 & 61 & 52 & - & 4.78 & 68 & 58 & 51 \\
                      \midrule
\multicolumn{1}{l|}{\multirow{4}{*}{\uppercase\expandafter{\romannumeral2}}} &
  InstructNav~\cite{long2024instructnav} & \ding{51} &
  - &
  - &
  - &
  - &
  - &
  6.74 &
  6.04 &
  42 &
  30 &
  22 &
  - &
  - &
  - &
  - &
  - \\
\multicolumn{1}{l|}{} & NaVid~\cite{zhang2024navid}    & \ding{55}  & 10.27 & 5.47 & 52 & 43 & 38  & 7.63  & 5.47 & 49  & 37 & 35  & -     & -    & -   & -  & -   \\
\multicolumn{1}{l|}{} & Uni-NaVid$^\dagger$~\cite{zhang2024uni} & \ding{55} & 10.04     & 4.45    & 65    & 58  & 53   & 9.71  & 5.58 & 53  & 47 & 43  & -     & -    & -   & -  & -   \\
\multicolumn{1}{l|}{} & NaVILA~\cite{cheng2024navila}  &  \ding{55}  & -     & -    & -    & -  & -   & -     & 5.37 & 58  & 50 & 46  & -     & -    & -   & -  & -   \\
\multicolumn{1}{l|}{} & NaVILA$^\dagger$~\cite{cheng2024navila}  &  \ding{55}  & -    & 4.82    & 66    & 58  & 53   & -     & 5.22 & 63  & 54 & 49  & -     & -    & -   & -  & -   \\ 
\multicolumn{1}{l|}{} & StreamVLN~\cite{wei2025streamvln}  &  \ding{55}  & -     & -    & -    & -  & -   & -     & 5.43 & 63  & 53 & 47  & -     & -    & -   & -  & -   \\ 
\multicolumn{1}{l|}{} & StreamVLN$^\dagger$~\cite{wei2025streamvln}  &  \ding{55}  & -     & 4.53    & 68    & 62  & 58   & -     & 4.98 & 64  & 57 & 52  & -     & -    & -   & -  & -   \\ 
\midrule
\uppercase\expandafter{\romannumeral3} &
  \textbf{CLASH (Ours)} & \ding{51} &
  12.40 &
  \textbf{3.20} &
  \textbf{80} &
  \textbf{73} &
  \textbf{65} &
  12.72 &
  \textbf{4.06} &
  \textbf{73} &
  \textbf{65} &
  \textbf{55} &
  13.06 &
  \textbf{4.10} &
  \textbf{74} &
  \textbf{66} &
  \textbf{58} \\ \bottomrule
\end{tabular}}
\end{table*}

\subsubsection{Metrics}
Trajectory Length (TL) presents the length (m) of the navigation trajectory. Navigation Error (NE) calculates the distance (m) between the predicted and actual stop locations. Success Rate (SR) gauges how often the predicted stop location is within a predefined distance from the true location. Oracle Success Rate (OSR) determines the frequency with which any point on the predicted path is within a certain distance of the goal. Success Rate weighted by Inverse Path Length (SPL) measures navigation effectiveness by combining success rate with the length of the route.

\subsubsection{Implementation details}
Our VLN-CE implementation builds on ETPNav~\cite{an2024etpnav}. The RSMP model is initialized with ScaleVLN~\cite{wang2023scaling} fine-tuned weights, with CLIP-H/14~\cite{radford2021learning} for image features and learnable backdoor and frontdoor causal modules from GOAT~\cite{wang2024causal}. Depth information is incorporated via a pre-trained ResNet-50~\cite{wijmansdd2019ddppo}. Training in Habitat uses a batch size of 16 for 15K iterations with AdamW ($2 \times 10^{-5}$), only on the standard R2R training split, and the best checkpoint is selected based on SR and SPL on validation-unseen. Compared with other LLM-based fine-tuning strategies that require large multi-GPU clusters, our method trains efficiently on only one single NVIDIA L40 GPU. REVERIE-CE is fine-tuned from the best R2R-CE checkpoint with identical settings. The RLMR deploys the open-source Qwen2.5-VL-72B~\cite{Qwen2.5-VL} locally via VLLM~\cite{kwon2023efficient} on 4 A800 GPUs.

\subsection{Comparisons with State-of-the-Arts (SoTA)}
\subsubsection{Results on R2R-CE}
Table~\ref{tab_r2rce} compares CLASH with current SoTA methods on the R2R-CE dataset. For clarity, we categorize existing approaches into three types: \texttt{Type-\uppercase\expandafter{\romannumeral1}} includes conventional end-to-end task-specific models that are fully fine-tuned on the R2R-CE dataset. \texttt{Type-\uppercase\expandafter{\romannumeral2}} represents recent efforts to incorporate large models into VLN-CE. 
While these large-model approaches perform well in monocula settings, they still fall short of panoramic task-specific models, a trend also observed in discrete VLN benchmarks~\cite{zhou2024navgpt,zhou2024navgpt2}. \texttt{Type-\uppercase\expandafter{\romannumeral3}} refers to our proposed collaborative small-large framework, CLASH. 
It shows that CLASH outperforms existing methods across splits and metrics.
Particularly, on the test unseen split, CLASH achieves the \textbf{1-st rank} in the VLN-CE test leaderboard\footnote{\url{https://eval.ai/web/challenges/challenge-page/719/overview}}, surpassing g3D-LF~\cite{wang2025g3d} by a relative 13.79\% in SR, and 13.73\% in SPL, respectively. Compared to approaches using large vision-language models, CLASH demonstrates substantial gains in both seen and unseen environments. For instance, compared with NaVILA~\cite{cheng2024navila}, CLASH increases relative SR by 25.86\% on val-seen, and 20.37\% on val-unseen, respectively.  

\subsubsection{Results on REVERIE-CE}
Table~\ref{tab_reverie_ce} compares our proposed CLASH method with other SoTA methods on the REVERIE-CE dataset. We mapped the REVERIE coordinates from the discrete environment into the Habitat simulator, enabling data collection, training, and model evaluation. It shows that CLASH outperforms previous methods on all metrics on both val-seen and val-unseen splits.
For instance, it shows that on the val-unseen set, CLASH relatively outperforms g3d-LF~\cite{wang2025g3d} by 2.92\% in SR and 11.76\% in SPL, respectively. On the val-seen set, CLASH improves ETPNav~\cite{an2024etpnav} by relative 24.94\% SR and 24.93\% SPL, respectively. This has significantly demonstrated the effectiveness and robustness of our proposed method.

\begin{table}[t]
\caption{Comparison with other SoTA methods on the REVERIE-CE~\cite{qi2020reverie}.}
\centering
\label{tab_reverie_ce}
\resizebox{\linewidth}{!}{
\begin{tabular}{@{}l|c|cccc|cccc@{}}
\toprule
\multirow{2}{*}{Method} & \multirow{2}{*}{Pano} & \multicolumn{4}{c|}{Val Seen}             & \multicolumn{4}{c}{Val Unseen}             \\
                        &    & NE$\downarrow$   & OSR$\uparrow$  & SR$\uparrow$ & SPL$\uparrow$    & NE$\downarrow$   & OSR$\uparrow$  & SR$\uparrow$ & SPL$\uparrow$  \\ \midrule
Seq2Seq~\cite{krantz_beyond_2020}    & \ding{55}         & 9.96 & 22.8 & 13.7 & 10.9  & 10.10 & 18.6 & 9.6  & 7.9  \\
CMA-TF~\cite{li2022reve-ce}         & \ding{55}        & 9.83 & 23.2 & 14.2 & 11.9         & 10.27 & 16.8 & 8.8  & 7.3  \\
CMA-SF~\cite{li2022reve-ce}         &  \ding{55}          & 9.84 & 30.4 & 18.9 & 16.1     & 10.20 & 29.5 & 13.2 & 9.9  \\
InstructNav~\cite{long2024instructnav}      &  \ding{51}                 & -    & -      & -    & -             & 7.44  & 31.5 & 25.2 & 19.1 \\
NaVid~\cite{zhang2024navid}         &   \ding{55}                 & -    & -    & -     & -             & 6.74  & 36.3 & 26.6 & 20.8 \\
ETPNav~\cite{an2024etpnav}       &      \ding{51}         & 6.65 & 49.1    & 40.9    & 36.1      & 7.01  & 36.2    & 28.9    & 24.2    \\ 
g3D-LF~\cite{wang2025g3d}      &     \ding{51}     & -    & -    & -    & -         & \textbf{6.50}  & 41.6 & 34.4 & 23.8 \\
\midrule
\textbf{CLASH (Ours)} & \ding{51}  & \textbf{5.38} & \textbf{53.3} & \textbf{51.1} & \textbf{45.1} & 6.82 & \textbf{43.0} & \textbf{35.4} & \textbf{26.6} \\ \bottomrule
\end{tabular}}
\end{table}

\subsection{Quantitative Analysis}
\label{subsec:ablation}
\subsubsection{Effect of different components affecting the RSMP}
Table~\ref{tab_rsmp_scalevln} presents an ablation of key design choices, revealing several insights:
First, leveraging pretrained weights significantly boosts RSMP performance. Both ETPNav (EN)~\cite{an2024etpnav} and ScaleVLN (SV)~\cite{wang2023scaling} adopt DUET~\cite{chen2022think} as their backbone, but differ in data augmentation: EN uses PREVALENT~\cite{hao2020towards} (61 MP3D scenes), while SV incorporates an additional 800 scenes from HM3D. As seen in \#1 and \#5, initializing with SV weights greatly improves generalization on val-unseen. However, combining Habitat training with SV’s large dataset offers diminishing returns and even degrades val-seen performance (\#3 and \#4), likely due to distribution mismatch. Therefore, we fine-tune only on the standard training set.
Second, incorporating depth input improves navigation accuracy (\#4 and \#6), consistent with EN’s findings.
Third, EN removes the local branch from DUET, limiting its ability to exploit pretrained representations. Restoring the dual-branch structure improves cross-modal alignment and boosts performance across both seen and unseen splits (\#5 and \#6).

\begin{table}[tbp]
\caption{Comparison of components impacting RSMP performance. ``FT Set” indicates the fine-tuning dataset. ``DB” denotes the use of the dual-branch architecture.}
\centering
\label{tab_rsmp_scalevln}
\resizebox{\linewidth}{!}{
\begin{tabular}{@{}c|l|l|c|c|ccc|ccc@{}}
\toprule
\multirow{2}{*}{\#} &
  \multirow{2}{*}{Init.} &
  \multirow{2}{*}{\begin{tabular}[c]{@{}l@{}}FT Set\end{tabular}} &
  \multirow{2}{*}{Depth} &
  \multirow{2}{*}{DB} &
  \multicolumn{3}{c|}{R2R-CE Val Seen} &
  \multicolumn{3}{c}{R2R-CE Val Unseen} \\ 
  &                   &              &   &   & OSR   & SR    & SPL   & OSR   & SR    & SPL   \\ \midrule
1 & EN     & Train & \ding{51} & \ding{55} & 71.00 & 66.00 & 59.00 & 65.00 & 57.00 & 49.00 \\
2 & EN     & Train & \ding{51} & \ding{51} & 72.79   & 66.37  & 60.99  & 62.13  & 55.71  & 49.98  \\
3 & SV & SV     & \ding{55} & \ding{51} & 70.82 & 64.14 & 58.31 & 69.44 & 63.78 & 55.82 \\
4 & SV & Train & \ding{55} & \ding{51} & 74.71 & 69.32 & 62.45 & 70.24 & 63.66 & 55.11 \\
5 & SV & Train & \ding{51} & \ding{55} & 71.63 & 66.24 & 61.66 & 66.10 & 60.07 & 54.71 \\
6 &
  SV &
  Train &
  \ding{51} &
  \ding{51} &
  \textbf{76.64} &
  \textbf{72.01} &
  \textbf{65.63} &
  \textbf{70.24} &
  \textbf{64.36} &
  \textbf{56.85} \\ \bottomrule
\end{tabular}}
\end{table}

\subsubsection{Effect of causal learning in RSMP} 
We integrate the causal learning (CL) modules~\cite{wang2024causal,wang2024causality} into the RSMP backbone during VLN-CE fine-tuning to improve generalization. As shown in Table~\ref{tab_rsmp_causal_learning}, on both R2R-CE and REVERIE-CE, CL improves SPL on both seen and unseen environments. For example, on the REVERIE-CE val-unseen split, SR increases from 30.79 to 35.65 and SPL from 26.41 to 28.46. 
However, on R2R, the gains are less pronounced, likely because the pretrained SV weights already generalize well and VLN-CE lacks a speaker model~\cite{wang2023pasts, wang2023res} to complement instruction diversity. This observation motivates our future work.
\begin{table}[t]
\caption{Effect of causal learning (CL) on RSMP.}
\centering
\label{tab_rsmp_causal_learning}
\resizebox{\linewidth}{!}{
\begin{tabular}{@{}c|l|c|ccc|ccc@{}}
\toprule
\multirow{2}{*}{\#} &
  \multirow{2}{*}{Dataset} &
  \multirow{2}{*}{w/ CL} &
  \multicolumn{3}{c|}{Val Seen} &
  \multicolumn{3}{c}{Val Unseen} \\
  &                          &   & OSR            & SR             & SPL            & OSR            & SR             & SPL            \\ \midrule
1 & \multirow{2}{*}{R2R-CE}     & \ding{55} & 76.64          & \textbf{72.01}          & 65.63          & 70.24          & 64.36          & 56.85          \\
2 &
   &
  \ding{51}  &
  \textbf{78.05} &
  71.25 &
  \textbf{65.90} &
  \textbf{71.00} &
  \textbf{64.53} &
  \textbf{57.86} \\ \midrule
3 & \multirow{2}{*}{REVERIE-CE} & \ding{55}  & 51.72          & 48.77          & 44.20 & 37.09          & 30.79          & 26.41          \\
4 &                          & \ding{51}  & \textbf{53.90} & \textbf{50.10} & \textbf{44.55}         & \textbf{42.49} & \textbf{35.65} & \textbf{28.46} \\ \bottomrule
\end{tabular}}
\end{table}

\subsubsection{Effect of the collaborative large-small hierarchy}
\begin{figure}[htbp]
    \centering
    \begin{minipage}{\linewidth}
        \centering
        \includegraphics[width=0.8\linewidth]{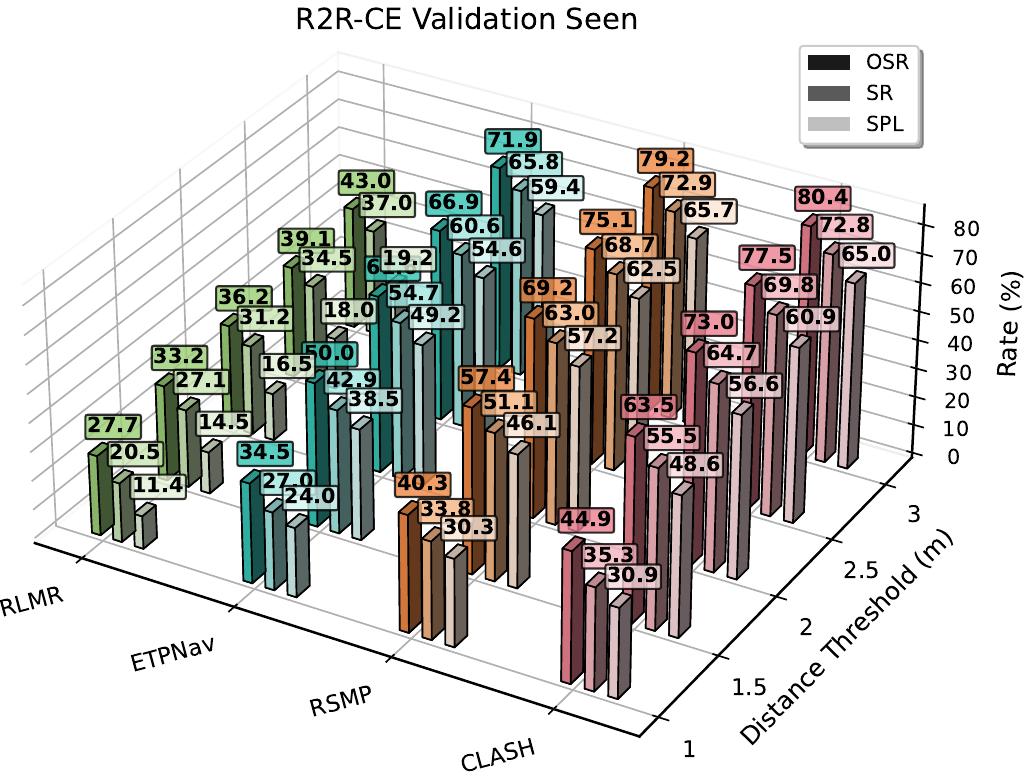}
    \end{minipage}
    
    \vspace{10pt}
    
    \begin{minipage}{\linewidth}
        \centering
        \includegraphics[width=0.8\linewidth]{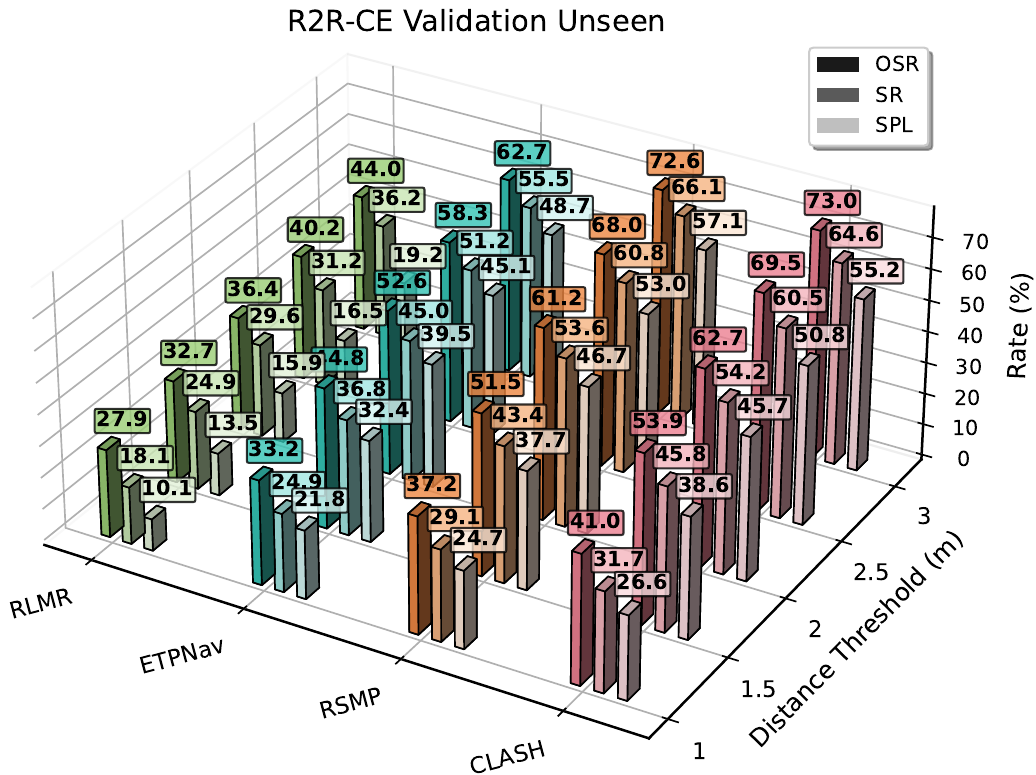}
    \end{minipage}
    \caption{Evaluation results under different success distance thresholds.}
    \label{fig:comparison_success_distance}
\end{figure}

Fig.~\ref{fig:comparison_success_distance} compares the performance of ETPNav, CLASH, and its single-model variants, namely the large-only model (RLMR) and the small-only model (RSMP), under different success distance thresholds (SDT). SDT is the maximum distance (in meters) allowed between the agent’s stop point and the goal for success; smaller values impose stricter criteria, with 3 meters as the default.
CLASH consistently outperforms ETPNav under all thresholds, with notable relative gains at SDT=3: +27.3\% SR / +18.4\% SPL (val-unseen) and +30.7\% SR / +28.8\% SPL (val-seen). At SDT=1, CLASH surpasses RSMP and RLMR by relative gains of 8.9\% and 75\% SR on val-unseen, indicating better goal proximity, a critical factor for downstream tasks like mobile manipulation. Similar trends are observed under stricter SDTs (1.0-2.5m), confirming CLASH’s robustness. However, we also observed that at SDT=3, CLASH occasionally shows slight drops in SR and SPL compared to the strong RSMP. This indicates that RSMP generalizes well in certain simulated environments and instruction styles. In contrast, RLMR may be affected by ambiguous instructions or spatial reasoning errors.

\subsubsection{Effect of the UCM}
Tab.~\ref{tab_rlmr_combination} compares different strategies for integrating RSMP and RLMR. Directly using RLMR for final decisions (\#1) yields the weak performance, and introducing an independent large-model arbiter (\#2) provides some gains by selecting between RSMP and RLMR outputs. \newcontent{Rows \#3–\#4 report results of the UCM using entropy and MC-dropout–based uncertainty estimation, both of which underperform compared with CP-based estimation.}
Rows \#5–\#8 evaluate our CP-based UCM under different error tolerances. The threshold $\tau$ determines how many candidates are forwarded to RLMR: a smaller $\tau$ invokes the large model more frequently. Using 50\% of the training data, we estimate $\tau$ corresponding to error tolerances $\epsilon \in {0.05\%, 0.1\%, 0.15\%, 0.25\%}$ as $\tau = {0.99, 0.97, 0.95, 0.93}$.
Overall, CP-based UCM consistently outperforms all other integration strategies. CLASH remains stable across $\tau$ values (std: 0.13 SR, 0.30 SPL), with $\tau = 0.97$ providing the better trade-off between seen and unseen environments.
\begin{table}[]
\caption{Collaboration Impact in Large-Small Hierarchy.}
\centering
\label{tab_rlmr_combination}
\resizebox{\linewidth}{!}{
\begin{tabular}{@{}l|l|ccc|ccc@{}}
\toprule
\multirow{2}{*}{\#} & \multirow{2}{*}{Method} & \multicolumn{3}{c|}{R2R-CE Val Seen}                    & \multicolumn{3}{c}{R2R-CE Val Unseen}          \\
  &                       & OSR   & SR    & SPL   & OSR   & SR    & SPL            \\ \midrule
1 & RLMR decides directly   & 76.86  & 68.12 & 56.73  & 72.43  & 62.59  & 50.51           \\
2 & Additional VLM decides & 79.08 & 71.76 & 61.86 & 71.87 & 63.98 & 53.92          \\ \midrule
\newcontent{3} & \newcontent{UCM (Entropy)}	& 79.85 & 71.63 & 63.39 & 72.52 & 64.04 & 54.52 \\
\newcontent{4} & \newcontent{UCM (MC Dropout)}	& 76.64 & 69.58 & 61.99 & 68.88 & 61.67 & 53.95 \\ \midrule
5                   & UCM (CP, $\tau=0.99$)        & 79.72          & 72.53          & 63.09          & \textbf{73.56} & \textbf{65.34} & 54.66 \\
6                   & UCM (CP, $\tau=0.97$)        & \textbf{80.36} & \textbf{72.79} & \textbf{64.99} & 73.01          & 64.64          & \textbf{55.18} \\
7 & UCM (CP, $\tau=0.95$)      & 79.33 & 71.89 & 64.55 & 73.34 & 64.31 & 55.08 \\
8 & UCM (CP, $\tau=0.93$)      & 80.23  & 72.53  & 64.85  & 72.14  & 64.44  & 55.01          \\ \bottomrule
\end{tabular}}
\end{table}

\subsubsection{Effect of point-goal controller}
\begin{figure}[t]
    \centering
    \includegraphics[width=\linewidth]{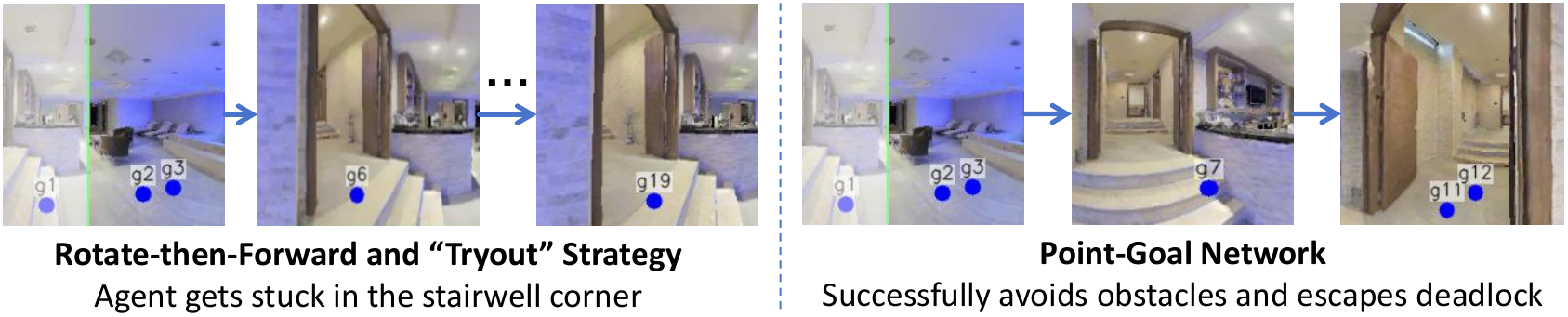}
    \caption{Comparison of different controllers in handling navigation deadlock.}
    \label{fig:deadlock_case}
\end{figure}

Table~\ref{tab_effect_controller} compares the rule-based Tryout controller from ETPNav with a fully learned DDPPO-based controller~\cite{wijmansdd2019ddppo}. Tryout often leads to deadlocks in cluttered scenes with elevation changes (Fig.~\ref{fig:deadlock_case}), whereas DDPPO provides more reliable obstacle avoidance and recovery. Replacing Tryout with DDPPO improves OSR by 2.29\% and 3.06\%, and SR by 1.25\% and 1.97\% on the val-seen and val-unseen splits. The slight reduction in SPL is due to DDPPO favoring safer, although sometimes longer, paths instead of the more aggressive shortcuts employed by Tryout.

\begin{table}[tbp]
  \caption{Effect of point-goal controller.}
  \centering
  \label{tab_effect_controller}
  \begin{tabular}{@{}l|c|ccc|ccc@{}}
  \toprule
  \multirow{2}{*}{\#} & \multirow{2}{*}{Controller} & \multicolumn{3}{c|}{R2R-CE Val Seen} & \multicolumn{3}{c}{R2R-CE Val Unseen} \\
    &        & OSR            & SR             & SPL            & OSR            & SR             & SPL            \\ \midrule
  1 & Tryout & 78.56          & 71.89          & \textbf{65.23} & 70.84          & 63.39          & \textbf{55.72} \\
  2 & DDPPO  & \textbf{80.36} & \textbf{72.79} & 64.99          & \textbf{73.01} & \textbf{64.64} & 55.18          \\ \bottomrule
  \end{tabular}
  \end{table}

\subsubsection{Effect of fine-tuning on RLMR}

\begin{figure}[t]
  \centering
  \includegraphics[width=\linewidth]{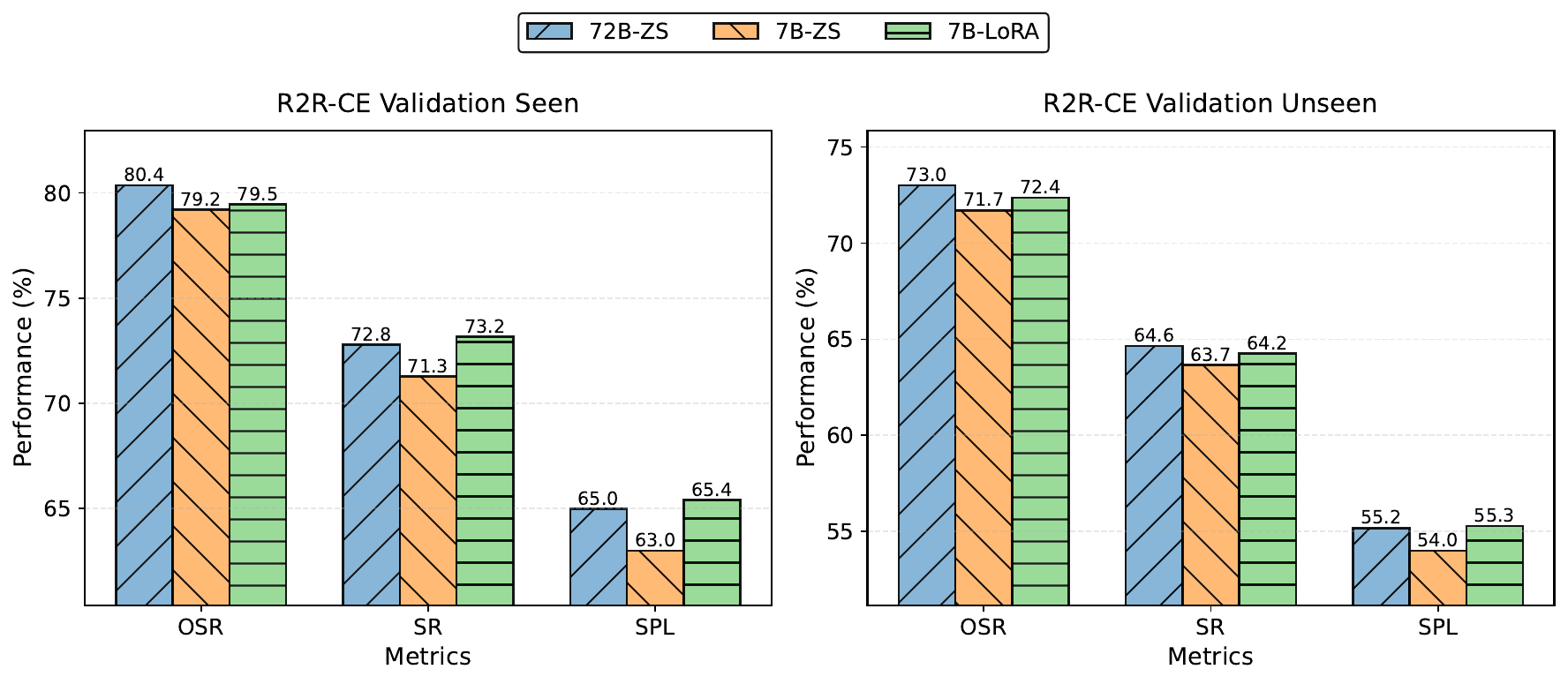}
  \caption{\newcontent{Effect of LoRA fine-tuning on the 7B model for CLASH.}}
  \label{fig:7b_bar_comparison}
\end{figure}

\newcontent{Given the high computational cost of the 72B model, we further investigate whether a smaller model can benefit from fine-tuning. We collect the 72B model’s correct responses on the training set and fine-tune the 7B model with LoRA (learning rate $2\times10^{-4}$, 12K iterations, one L40 GPU). As shown in Fig.~\ref{fig:7b_bar_comparison}, the fine-tuned 7B model achieves clear improvements over its zero-shot  (ZS) version and comes close to the 72B model in simulation. However, in real-world deployment (Sec.~\ref{subsec:real_world_results}), its performance drops significantly (72B-ZS SR 55\% \textit{vs.} 7B-Sft SR 35\% on average), suggesting dataset-specific overfitting and limited spatial reasoning capacity. Overall, while distillation reduces computation cost, the 72B zero-shot model remains more reliable in the real world. Advances in base MLLMs and more effective small-model adaptation may help narrow this gap.}

\subsubsection{\newcontent{Analysis on the model efficiency}}
\label{subsubsec:model_efficiency}
  
\newcontent{Tab.~\ref{tab:model_efficiency} reports a comprehensive comparison of computational efficiency across methods. Specifically, AET means the average episode time, and AST means the average step time. All baselines are evaluated using their official implementations on a single NVIDIA L40 GPU based on Habitat. The 72B model is deployed on four A800 GPUs with intra-node communication. The environment and batch size are both set to 1. For CLASH, the RSMP module is approximately 1B parameters, and RLMR includes the 72B-ZS model and the 7B LoRA-fine-tuned model.
The methods fall into two categories. Step-by-step action models (CMA, NaVID, UniNaVID, StreamVLN) have low per-step latency but require many steps per episode, resulting in longer overall episode times. In contrast, waypoint-based models (ETPNav, g3d-LF-Monocular, and OpenNav with Qwen2.5-VL-7B) predict multi-step waypoints, reducing the number of required steps and thus shortening episode durations.
The quantitative results indicate that CLASH achieves mid-level computational efficiency among the compared models while attaining the highest accuracy. When a fine-tuned 7B model is used, the efficiency can be further improved. This suggests that future work may explore applying the proposed framework to smaller models, aiming to retain the commonsense strengths of large models while achieving a more favorable balance between efficiency and accuracy.}
\begin{table}[t]
  \caption{\newcontent{Model efficiency comparison.}}
  \centering
  \label{tab:model_efficiency}
  \resizebox{\linewidth}{!}{
  \begin{tabular}{l|l|c|c|cc|ccc}
  \toprule
  Method & Params & AET(s)$\downarrow$ & AST(s)$\downarrow$ & \multicolumn{2}{c|}{Val Seen} & \multicolumn{2}{c}{Val Unseen} \\
   & & $(\pm \text{std})$ & $(\pm \text{std})$ & SR$\uparrow$ & SPL$\uparrow$ & SR$\uparrow$ & SPL$\uparrow$ \\
  \midrule
  CMA~\cite{krantz_beyond_2020} & 37M & $2.74 \pm 1.32$ & $0.03 \pm 0.09$ & 51 & 45 & 34 & 28 \\
  ETPNav~\cite{an2024etpnav} & 299M & $0.74 \pm 0.68$ & $0.07 \pm 0.19$ & 66 & 59 & 57 & 49 \\
  g3d-LF-Mono~\cite{wang2025g3d} & 763M & $12.94 \pm 5.08$ & $1.18 \pm 0.38$ & 47 & 35 & 49 & 37 \\
  OpenNav~\cite{qiao2024open} & 10B & $162.41 \pm 8.97$ & $27.00 \pm 3.64$ & - & - & 19 & 16 \\
  NaVID~\cite{zhang2024navid} & 7B & $21.08 \pm 15.12$ & $0.69 \pm 0.92$ & 43 & 38 & 37 & 35 \\
  UniNaVID~\cite{zhang2024uni} & 7B & $9.93 \pm 2.79$ & $0.18 \pm 0.02$ & 58 & 53 & 47 & 43 \\
  NaVILA~\cite{cheng2024navila} & 7B & $62.02 \pm 65.65$ & $0.86 \pm 0.17$ & 58 & 53 & 54 & 49 \\
  StreamVLN~\cite{wei2025streamvln} & 7B & $42.37 \pm 41.43$ & $0.71 \pm 0.16$ & 62 & 58 & 57 & 52 \\ \midrule
  CLASH (7B-LoRA) & 8B & $17.10 \pm 6.97$ & $1.92 \pm 1.30$ & 73 & 65 & 64 & 55 \\
  CLASH (72B-ZS) & 73B & $31.55 \pm 12.61$ & $3.63 \pm 2.66$ & 73 & 65 & 65 & 55 \\
  \bottomrule
  \end{tabular}}
  \end{table}

\subsection{Qualitative Analysis}
\label{subsec:visualization}

\subsubsection{Trajectory visualization}
\label{subsubsec:visualization_trajectories}
\begin{figure*}[htbp]
    \centering
    \includegraphics[width=0.95\linewidth]{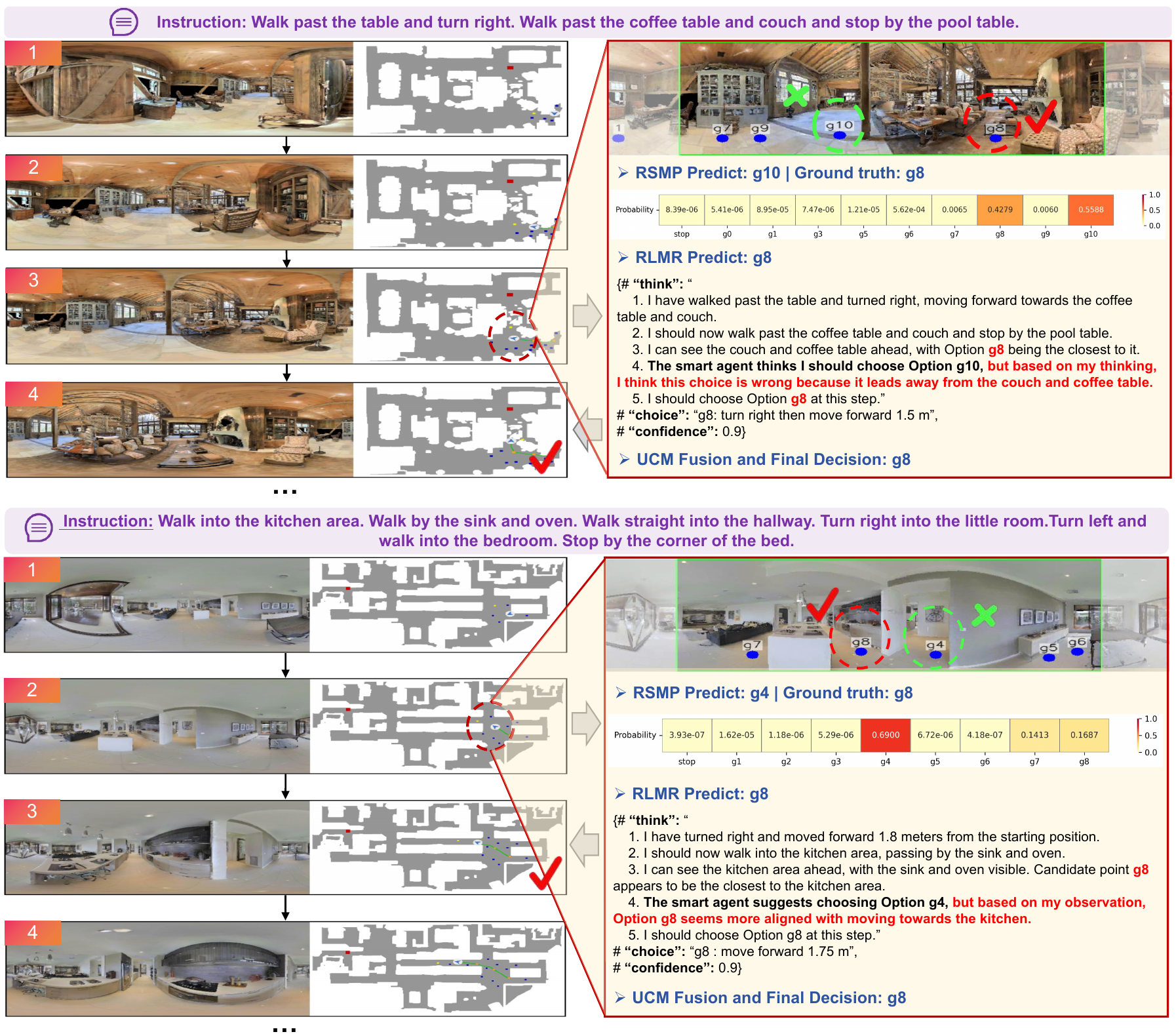}
    \caption{\newcontent{Visualization of navigation trajectories in VLN-CE.} At each step, the left panel shows the panoramic view, and the right panel displays a top-down map with the target (red), candidate waypoints (blue), and the robot's position and orientation (arrow).}
    \label{fig:traj_vis}
\end{figure*}

Fig.~\ref{fig:traj_vis} illustrates how CLASH operates in the VLN task and assists the agent in making more reliable decisions under ambiguous navigation scenarios. Take the first trajectory as an example, RSMP incorrectly proposes waypoint \texttt{g10}, which does not align with the implied actions \textit{``turn right”} and \textit{``walk past the coffee table.”} \newcontent{The heatmap of the probability distribution predicted by RSMP further shows its uncertainty: while \texttt{g10} receives the highest probability ($\textbf{p}^{\text{RSMP}}_{g10}=0.56$), \texttt{g8} is a strong alternative ($\textbf{p}^{\text{RSMP}}_{g8}=0.43$). With the strong commonse reasoning, RLMR detects this inconsistency and overrides the RSMP output with \texttt{g8}, which is consistent with the instruction. The final action, selected by UCM, allows the agent to successfully complete the task. A similar phenomenon occurs in the second trajectory example. When the agent should \textit{``walk into the kitchen area,''} RSMP incorrectly assigns the highest probability to \texttt{g4}. With RLMR and UCM, the correct node \texttt{g8} is identified, and the agent’s decision is corrected.}

\section{Real-world Experiments}
\label{sec_exp_sim2real}
\subsection{Experimental Settings}
Beyond proposing the novel CLASH framework, we develop practical low-level modules for real-world deployment (Sec.~\ref{sec_low_level_exe}). As shown in Fig.~\ref{fig_realDeploy_demo1}, we use a custom wheeled robot equipped with an Insta360 X4 panoramic camera (1.5m height) and a base-mounted 3D LiDAR. The robot communicates with a remote server via Wi-Fi for data processing and control.
Due to hardware constraints, the panoramic camera lacks aligned depth data, and LiDAR-camera calibration is imprecise. Thus, instead of using the pretrained waypoint predictor, we apply a simple LiDAR-based clustering algorithm to generate navigable waypoints.
Using ROS's TF tree, we retrieve both robot-relative and world coordinates for each waypoint, which are fed into the CLASH network for inference. The selected waypoint serves as the navigation target.
For execution, we use a SLAM-based local controller, which provides robust obstacle avoidance without requiring training or adaptation.

\subsection{Experimental Results}
\label{subsec:real_world_results}
\begin{figure}[htbp]
    \centering
    \includegraphics[width=\linewidth]{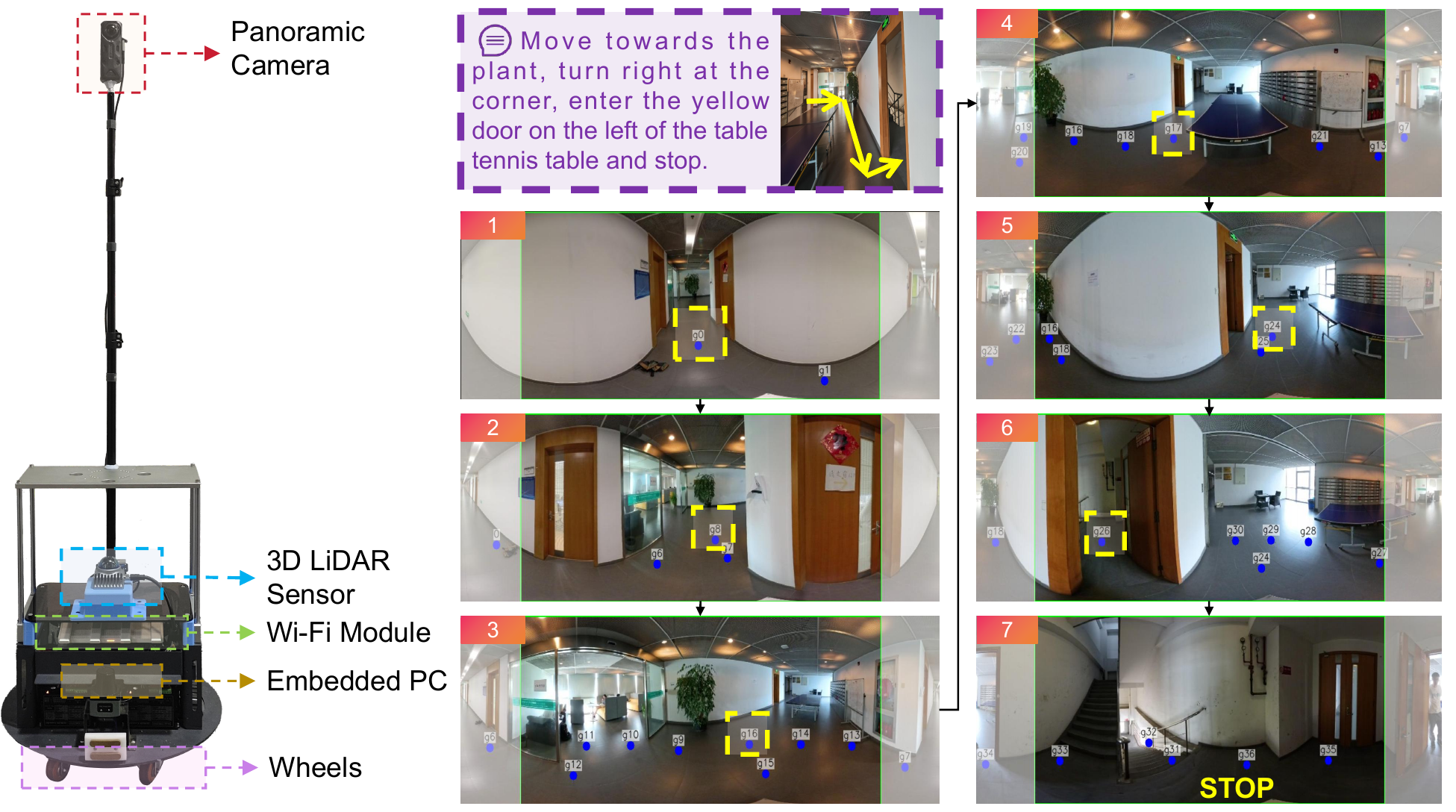}
    \caption{The navigation trajectory and observations in a real-world scenario using the wheeled robot. The yellow boxes mark selected waypoints.}
    \label{fig_realDeploy_demo1}
\end{figure}

As shown in Fig.~\ref{fig_realDeploy_demo1}, we conducted real-world experiments in university laboratories and classroom buildings. No prior maps were used, and the model was not fine-tuned on real-world scenes. As the depth cannot be directly obtained via the panoramic camera, we use the model trained without the depth as input. 
We tested a total of 40 episodes, categorized into three difficulty levels based on instruction complexity: easy (10 episodes), medium (14), and hard (16). Specifically, easy instructions are short and direct, such as \textit{“Go out of the door and wait.”} Medium instructions include multiple steps, e.g., \textit{“Go into the room in front of you, turn right, and stop near the girl sitting on the chair.”} Hard instructions involve longer navigation paths, e.g., \textit{“Turn around and go out of the room. Then turn right, walk down the hallway, and turn left at the intersection. Stop in front of the first elevator door.”} 

\begin{figure}[t]
  \centering
  \includegraphics[width=0.6\linewidth]{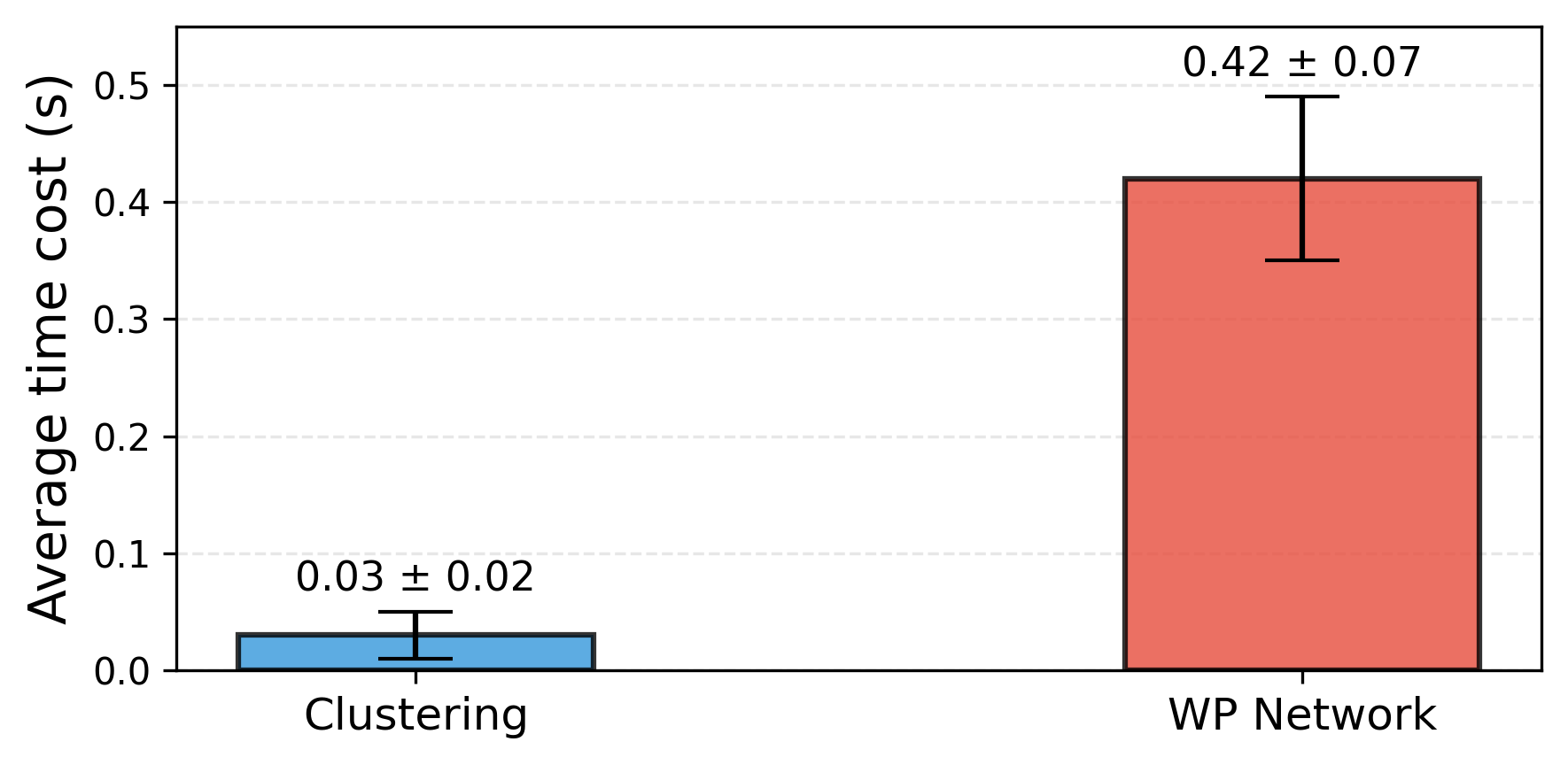}
  \caption{\newcontent{Time cost comparison of different waypoint generation methods.}}
  \label{fig:waypoint_time_comparison}
\end{figure}

\newcontent{In Fig.~\ref{fig:waypoint_time_comparison}, we compare the computational cost of different waypoint generation methods on a robot equipped with an NVIDIA Jetson Orin Nano Developer Kit. Since pixel-aligned panoramic depth is unavailable from the real-world panoramic camera, simulated data are used for evaluating the waypoint predictor (WP) network. The results indicate that on resource-constrained edge devices, the proposed KMeans-based clustering method generates candidate reachable points in only 0.03s, compared to 0.42s for the WP network, achieving an approximate 14-fold speedup.}

\begin{figure}[t]
    \centering
    \includegraphics[width=\linewidth]{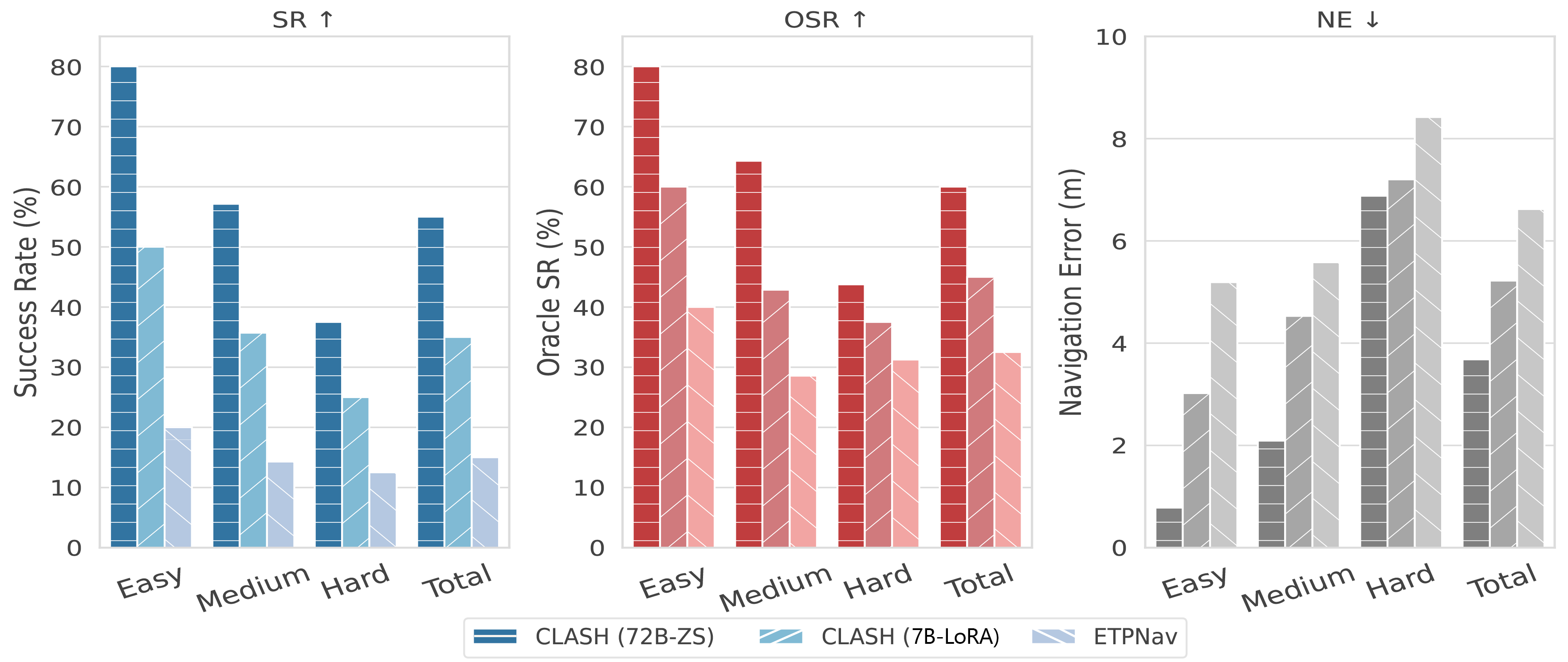}
    \caption{\newcontent{Quantitative comparison results of the real-world validation.}}
    \label{fig_real_world_etpnav}
\end{figure}

Quantitative results are shown in Fig.~\ref{fig_real_world_etpnav}. Due to the large domain gap between MP3D and our office-style real-world environments, ETPNav generalizes poorly. In contrast, CLASH, combining a strong RSMP with a generalized RLMR capable of broader perception and reasoning, achieves substantially better performance and robustness in real-world deployment. \newcontent{We further compare using 72B-ZS and 7B-LoRA as RLMR's backbone. The results indicate that the 72B-ZS model retains stronger commonsense reasoning in real environments than the 7B-LoRA model. Addressing the latency introduced by large models remains an important direction for future work.}

\section{Conclusion}
This paper presents CLASH, a novel collaborative framework that integrates reactive planning from a task-specific small model with reflective reasoning from a vision-language large model, enabling robust and interpretable vision-and-language navigation. Through uncertainty-aware decision fusion, panoramic prompting, and practical low-level control, CLASH achieves SoTA performance on VLN-CE and demonstrates strong real-world generalization without in-domain fine-tuning. Our findings highlight the importance of object-level visual grounding, embodied spatial perception, and model collaboration. We hope CLASH provides a foundation for future research in bridging general-purpose intelligence with embodied decision-making in complex environments.

\bibliographystyle{IEEEtran}
\bibliography{main}

\newpage
\appendices

\section{Supplementary for CLASH}
\label{appendix_sec_vlm_prompt}

\subsection{Limitations and Failure Case Analysis}
While CLASH achieves state-of-the-art performance on VLN-CE benchmarks, incorporating a large vision–language model inevitably increases computational cost. As shown in Tab.~\ref{tab:inference_time}, inference with the 72B model is approximately 30 times slower per step than with the 1B model, which poses challenges for real-time deployment. The proposed UCM mechanism alleviates this issue by invoking the RLMR only when the RSMP exhibits low confidence, thereby reducing unnecessary large-model calls and improving overall decision-making efficiency.

\begin{table}[htbp]
\centering
\caption{Model Size and Inference Time Comparison}
\begin{tabular}{lcc}
\toprule
\textbf{Module} & \textbf{Size (B)} & \textbf{Inference Time (s)} \\
\midrule
RSMP  & 1  & $0.225 \pm 0.193$ \\
RLMR  & 7  & $3.061 \pm 1.085$ \\
RLMR  & 72   & $6.897 \pm 1.034$ \\
\bottomrule
\end{tabular}
\label{tab:inference_time}
\end{table}

\begin{figure}[h]
    \centering
    \includegraphics[width=0.8\linewidth]{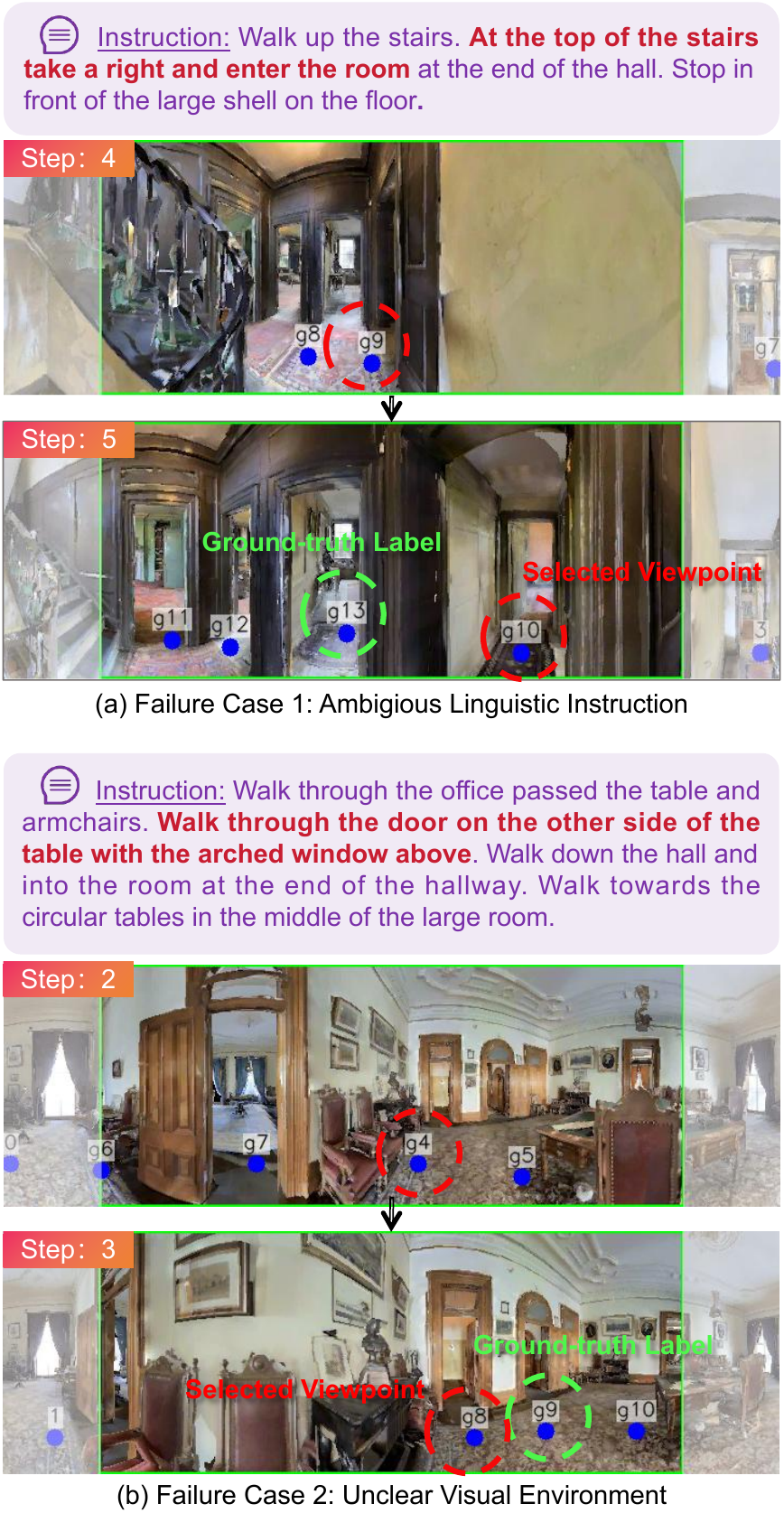}
    \caption{Examples of two typical failure cases. The red bold text highlights the sub-instruction encountered when the agent made an incorrect decision.}
    \label{fig_failure_case}
\end{figure}

Fig.\ref{fig_failure_case} illustrates two representative failure cases. In the first case, the failure results from ambiguity in the instruction—e.g., \textit{``take a right and enter the room at the end of the hall."}—where multiple candidate rooms are located on the right, and the agent incorrectly selects the farthest one instead of the intended second room. Even humans may struggle with such ambiguous references in the absence of further context. In the second case, the failure occurs because the agent is unable to accurately ground the target door described as \textit{``with the arched window above."} This highlights the remaining challenge of fine-grained vision-language alignment.

In addition, real-world experiments reveal that both large and small models are relatively insensitive to directional cues like \textit{“turn left”} or \textit{“turn right,”} but respond more reliably to semantically meaningful landmarks such as \textit{“door,”} \textit{“plant,”} or \textit{“table.”} This suggests that current VLN models align language and vision primarily at the object level, with limited embodied spatial understanding. Enhancing instructions or environments with salient, spatially anchored landmarks can thus improve navigation success.

Future work will focus on three primary directions: (1) improving the inference efficiency of large models through lightweight adaptation or distillation; (2) enabling agents to conduct interactive dialogue to clarify ambiguous instructions; and (3) enhancing the visual realism of simulation environments to further narrow the sim-to-real gap.

\subsection{Visualized Navigation Trajectories}
\begin{figure*}[t]
    \centering
    \includegraphics[width=0.85\linewidth]{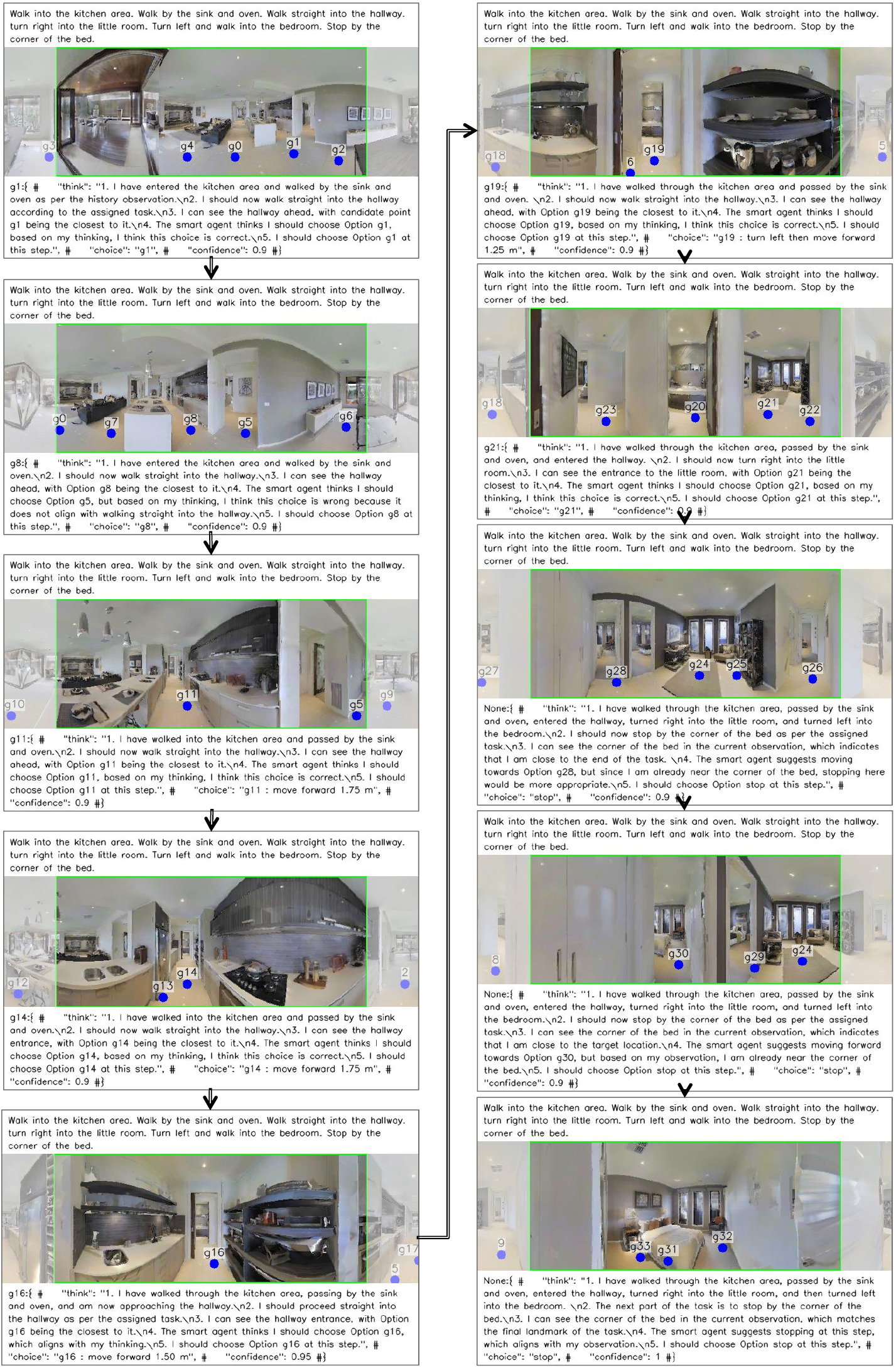}
    \caption{Visualization of episode ID 40 on the val-unseen split. The collaboration mechanism is evident in this trajectory. For example, in Step 2, RLMR corrected the decision to move from \texttt{g5} to \texttt{g8}. Near the stopping point, RSMP took the lead, overriding RLMR’s influence to ensure correct termination.}
    \label{fig_appendix_simulation1}
\end{figure*}

\begin{figure*}[t]
    \centering
    \includegraphics[width=0.85\linewidth]{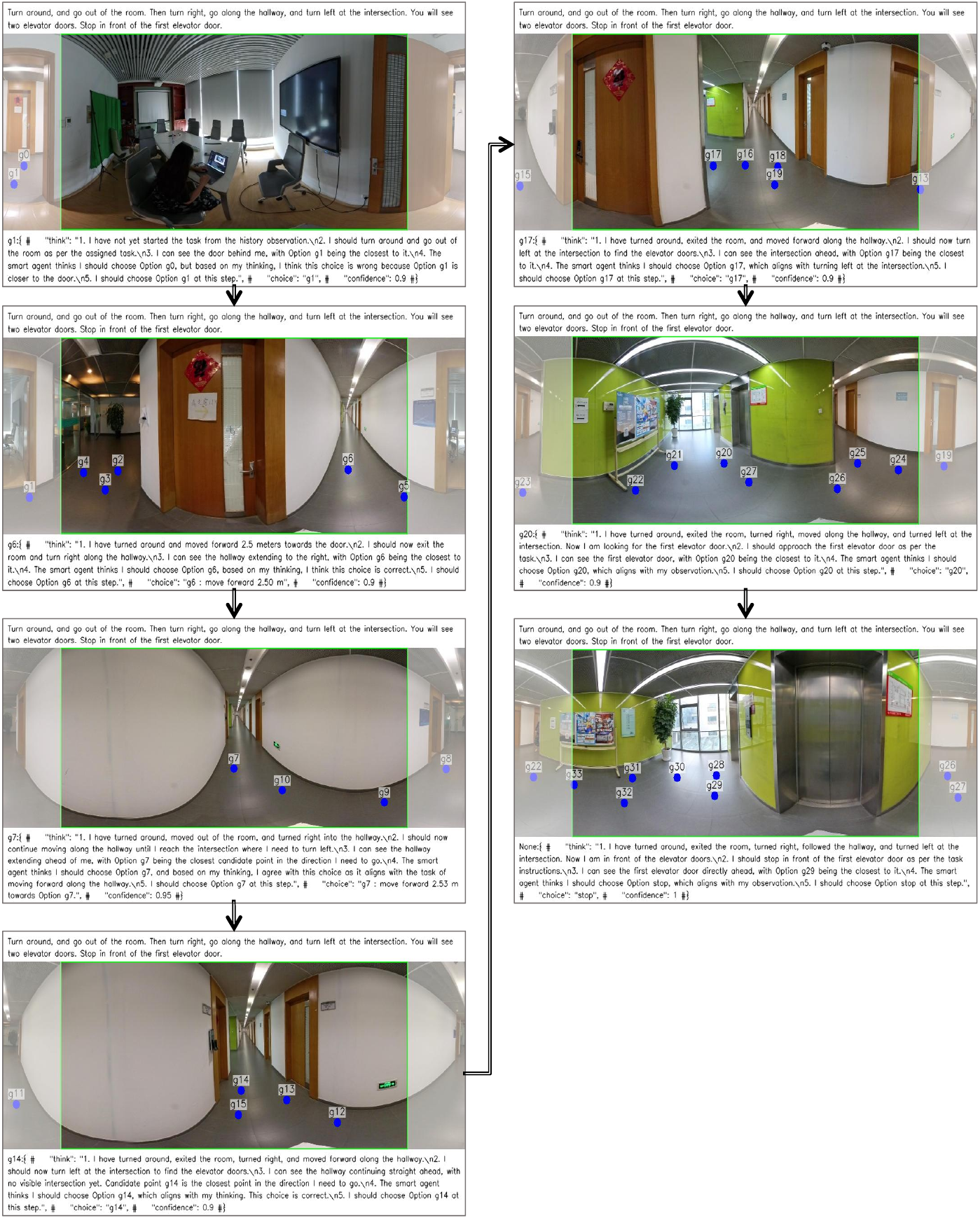}
    \caption{Visualization of episode in the real-world scenarios. The given instruction is shown at the top of each frame.}
    \label{fig_appendix_realworld1}
\end{figure*}

Fig.\ref{fig_appendix_simulation1} and Fig.\ref{fig_appendix_realworld1} present complete navigation trajectories in both simulation and real-world environments, along with RLMR’s step-by-step responses. The given instruction is displayed at the top of each frame, while RLMR’s reasoning outputs are shown at the bottom. Thanks to the chain-of-thought process, the agent’s decision-making becomes more interpretable and transparent, allowing for clearer analysis of its behavioral patterns and output tendencies.

\begin{figure*}[t]
    \centering
    \includegraphics[width=0.85\linewidth]{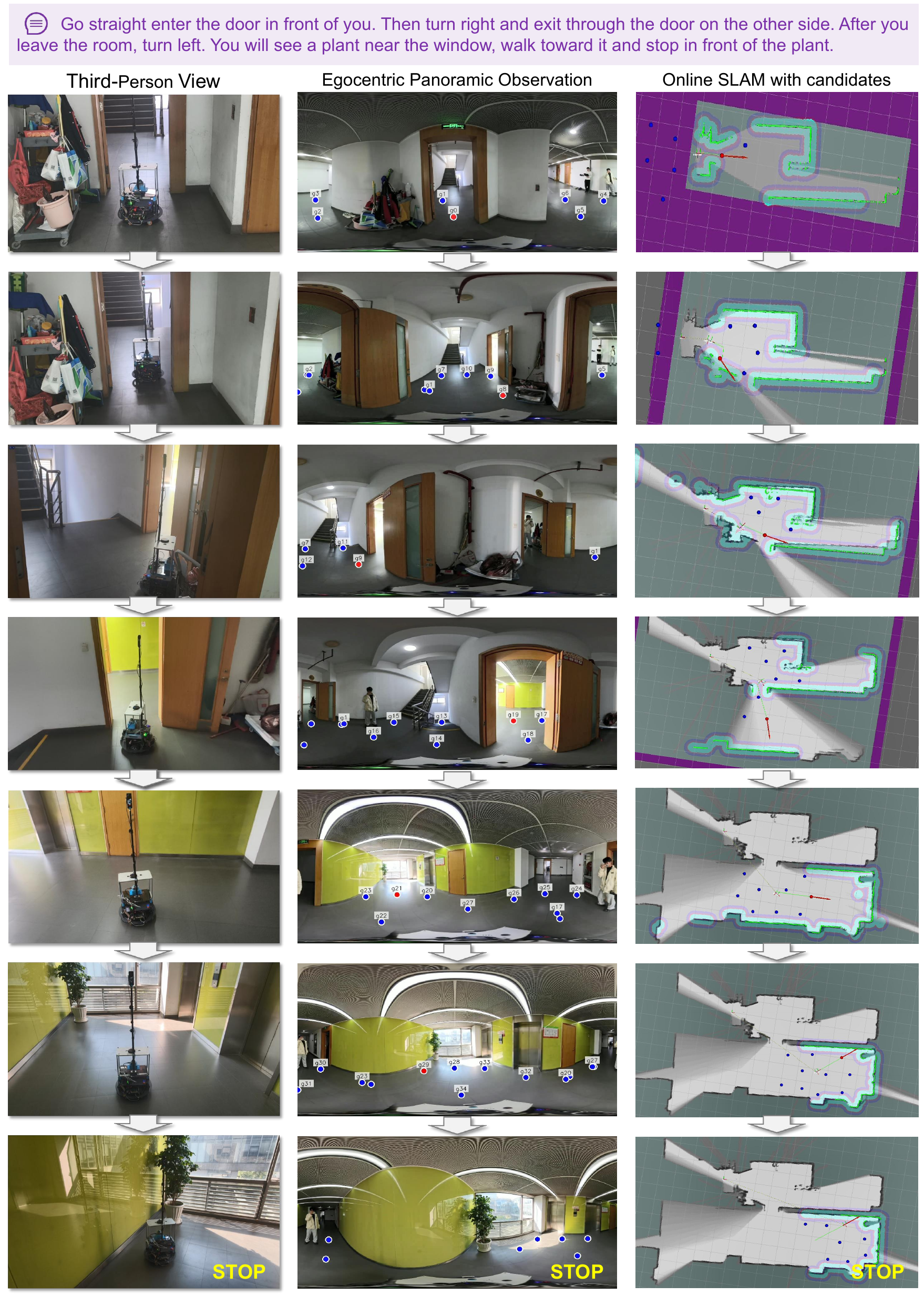}
    \caption{Visualization of a real-world episode. At each step, we show the third-person view, the egocentric panoramic observation, and the online SLAM map with candidate waypoints. The red dot indicates the selected waypoint.}
    \label{fig_appendix_realworld2}
\end{figure*}

\begin{figure*}[t]
   \centering
   \includegraphics[width=0.85\linewidth]{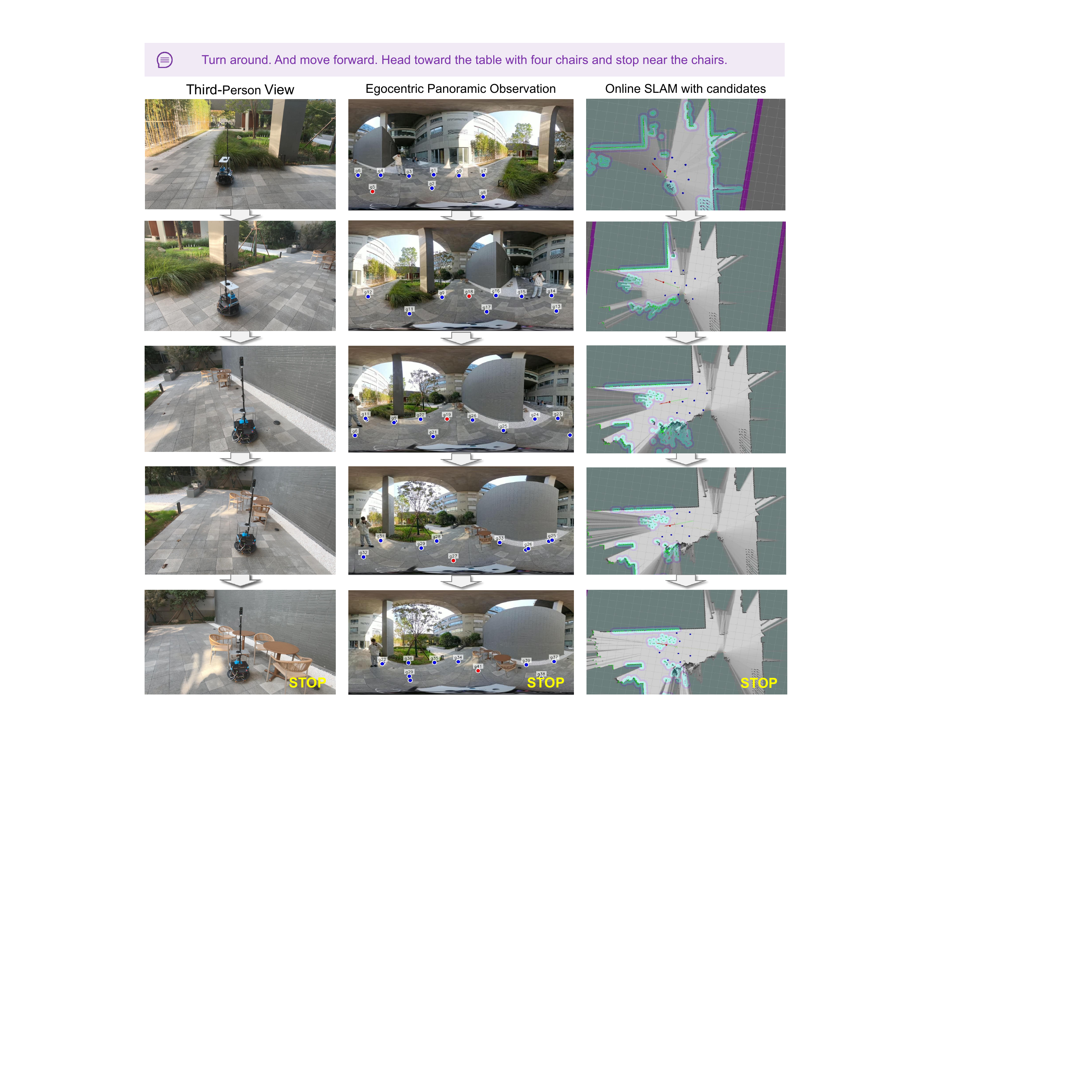}
   \caption{Visualization of a real-world episode. At each step, we show the third-person view, the egocentric panoramic observation, and the online SLAM map with candidate waypoints. The red dot indicates the selected waypoint.}
   \label{fig_appendix_realworld3}
\end{figure*}

\vfill

\end{document}